\documentclass[10pt,twocolumn,letterpaper]{article}

\usepackage{iccv}
\usepackage{times}
\usepackage{epsfig}
\usepackage{graphicx}
\usepackage{color}
\usepackage{amsmath}
\usepackage{amssymb}
\usepackage{mathtools}
\usepackage{booktabs}
\usepackage{multirow}
\usepackage{makecell}
\usepackage[accsupp]{axessibility}
\newcommand{\tabincell}[2]{\begin{tabular}{@{}#1@{}}#2\end{tabular}}

\DeclarePairedDelimiter\norm{\lVert}{\rVert_2}
\DeclarePairedDelimiter\abs{|}{|}

\usepackage[pagebackref=true,breaklinks=true,colorlinks,bookmarks=false]{hyperref}

\iccvfinalcopy 

\def\iccvPaperID{****} 

\ificcvfinal\fi

\begin{document}

\title{OMNet: Learning Overlapping Mask for Partial-to-Partial \\Point Cloud Registration}

\author{
Hao Xu$^{1,2}$ \hspace{0.5cm} 
Shuaicheng Liu$^{1,2*}$ \hspace{0.5cm} 
Guangfu Wang$^2$ \hspace{0.5cm} 
Guanghui Liu$^1$\thanks{Corresponding author} \hspace{0.5cm} 
Bing Zeng$^1$ \\
\\
\textsuperscript{1}University of Electronic Science and Technology of China \\
\textsuperscript{2}Megvii Technology \\
}

\maketitle
\ificcvfinal\thispagestyle{empty}\fi

\begin{abstract}
Point cloud registration is a key task in many computational fields. Previous correspondence matching based methods require the inputs to have distinctive geometric structures to fit a 3D rigid transformation according to point-wise sparse feature matches. However, the accuracy of transformation heavily relies on the quality of extracted features, which are prone to errors with respect to partiality and noise. In addition, they can not utilize the geometric knowledge of all the overlapping regions. On the other hand, previous global feature based approaches can utilize the entire point cloud for the registration, however they ignore the negative effect of non-overlapping points when aggregating global features. In this paper, we present OMNet, a global feature based iterative network for partial-to-partial point cloud registration. We learn overlapping masks to reject non-overlapping regions, which converts the partial-to-partial registration to the registration of the same shape. Moreover, the previously used data is sampled only once from the CAD models for each object, resulting in the same point clouds for the source and reference. We propose a more practical manner of data generation where a CAD model is sampled twice for the source and reference, avoiding the previously prevalent over-fitting issue. Experimental results show that our method achieves state-of-the-art performance compared to traditional and deep learning based methods. Code is available at \href{https://github.com/megvii-research/OMNet}{https://github.com/megvii-research/OMNet}.
\end{abstract}

\vspace{-0.4cm}
\section{Introduction}
\begin{figure}[t]
    \centering
    \includegraphics[width=\linewidth]{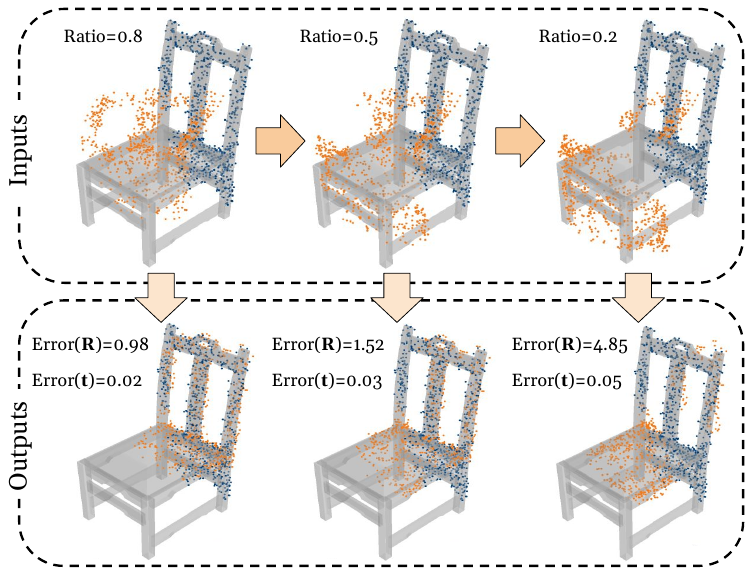}\\
    \caption{Our OMNet shows robustness to the various overlapping ratios of inputs. All inputs are transformed by the same 3D rigid transformation. $\operatorname{Error}(\mathbf{R})$ and $\operatorname{Error}(\mathbf{t})$ are isotropic errors.} 
    \vspace{-0.37cm}
    \label{fig:teaser}
\end{figure}
Point cloud registration is a fundamental task that has been widely used in various computational fields, e.g., augmented reality~\cite{azuma1997survey, carmigniani2011augmented, billinghurst2015survey}, 3D reconstruction~\cite{izadi2011kinectfusion, merickel19883d} and autonomous driving~\cite{yurtsever2020survey, geiger2012we}. It aims to predict a 3D rigid transformation aligning two point clouds, which may be potentially obscured by partiality and contaminated by noise.

Iterative Closest Point (ICP)~\cite{besl1992method} is a well-known algorithm for the registration problem, where 3D transformations are estimated iteratively by singular value decomposition (SVD) given the correspondences obtained by the closest point search.
However, ICP easily converges to local minima because of the non-convexity problem.
For this reason, many methods~\cite{rusinkiewicz-normal-sampling, fitzgibbon2003robust, segal2009generalized, bouaziz2013sparse, pomerleau2015review, yang2013go} are proposed to improve the matching or search larger transformation space, and one prominent work is the Go-ICP~\cite{yang2013go}, which uses a branch-and-bound algorithm to cross the local minima. Unfortunately, it is much slower than the original ICP. All these methods are sensitive to the initial positions.

Recently, several deep learning (DL) based approaches are proposed~\cite{wang2019deep,aoki2019pointnetlk, wang2019prnet, sarode2019pcrnet, yew2020-RPMNet, huang2020feature, idam, yuan2020deepgmr} to handle the large rotation angles. Roughly, they could be divided into two categories: correspondence matching based methods and global feature based methods. Deep Closest Point (DCP)~\cite{wang2019deep} determines the correspondences from learned features. DeepGMR~\cite{yuan2020deepgmr} integrates Gaussian Mixture Model (GMM) to learn pose-invariant point-to-GMM correspondences. However, they do not take the partiality of inputs into consideration. PRNet~\cite{wang2019prnet}, RPMNet~\cite{yew2020-RPMNet} and IDAM~\cite{idam} are presented to mitigate this problem by using Gumbel–Softmax~\cite{jang2016categorical} with Sinkhorn normalization~\cite{sinkhorn1964relationship} or a convolutional neural network (CNN) to calculate matching matrix. However, these methods require the inputs to have distinctive local geometric structures to extract reliable sparse 3D feature points. As a result, they can not utilize the geometric knowledge of the entire overlapping point clouds. In contrast, global feature based methods overcome this issue by aggregating global features before estimating transformations, e.g., PointNetLK~\cite{aoki2019pointnetlk}, PCRNet~\cite{sarode2019pcrnet} and Feature-metric Registration (FMR)~\cite{huang2020feature}. However, all of them ignore the negative effect of non-overlapping regions.

In this paper, we propose OMNet: an end-to-end iterative network that estimates 3D rigid transformations in a coarse-to-fine manner while preserving robustness to noise and partiality. To avoid the negative effect of non-overlapping points, we predict overlapping masks for the two inputs separately at each iteration. Given the accurate overlapping masks, the non-overlapping points are rejected during the aggregation of global features, which converts the partial-to-partial registration to the registration of the same shape. As such, regressing rigid transformation becomes easier given global features without interferes. This desensitizes the initial position of the inputs and enhances the robustness to noise and partiality. Fig.~\ref{fig:teaser} shows the robustness of our method to the inputs with different overlapping ratios. Experiments show that our approach achieves state-of-the-art performance compared with previous algorithms.

Furthermore, ModelNet40~\cite{wu20153d} dataset is adopted for the registration~\cite{wang2019deep,aoki2019pointnetlk, wang2019prnet, sarode2019pcrnet, yew2020-RPMNet, huang2020feature, idam, yuan2020deepgmr}, which has been originally applied to the task of classification and segmentation. Previous works follow the data processing of PointNet~\cite{qi2017pointnet}, which has two problems: (1) a CAD model is sampled only once during the point cloud generation, yielding the same source and the reference points, which often causes over-fitting issues; (2) ModelNet40 dataset involves some axisymmetrical categories where it is reasonable to obtain an arbitrary angle on the symmetrical axis. We propose a more suitable method to generate a pair of point clouds. Specifically, the source and reference point clouds are randomly sampled from the CAD model separately. Meanwhile, the data of axisymmetrical categories are removed. In summary, our main contributions are:

\begin{itemize}
\setlength{\itemsep}{0pt}
\setlength{\parsep}{0pt}
\setlength{\parskip}{0pt}
\item We propose a global feature based registration network OMNet, which is robust to noise and different partial manners by learning masks to reject non-overlapping regions. Mask prediction and transformation estimation can be mutually reinforced  during iteration.
\item We expose the over-fitting issue and the axisymmetrical categories that existed in the ModelNet40 dataset when adopted to the registration task. In addition, we propose a more suitable method to generate data pairs for the registration task.
\item We provide qualitative and quantitative comparisons with other works under clean, noisy and different partial datasets, showing state-of-the-art performance.
\end{itemize}

\section{Related Works}
\paragraph{Correspondence Matching based Methods.}
Most correspondence matching based methods solve the registration problem by alternating two steps: (1) set up correspondences between the source and reference point clouds; (2) compute the least-squares rigid transformation between the correspondences. ICP~\cite{besl1992method} estimates correspondences using spatial distances. Subsequent variants of ICP improve performance by detecting keypoints~\cite{gelfand2003geometrically, rusinkiewicz-normal-sampling} or weighting correspondences~\cite{godin1994three}. However, due to the non-convexity of the first step, they are often strapped into local minima. To address this, Go-ICP~\cite{yang2013go} uses a branch-and-bound strategy to search the transformation space at the cost of a much slower speed. Recently proposed Symmetric ICP~\cite{rusinkiewicz2019symmetric} improves the original ICP by designing the objective function. Instead of using spatial distances, PFH~\cite{rusu2008aligning} and FPFH~\cite{FPFH} design rotation-invariant descriptors and establish correspondences from handcrafted features. To avoid computation of RANSAC~\cite{fischler1981random} and nearest-neighbors, FGR~\cite{zhou2016fast} uses alternating optimization techniques to accelerate iterations. 

More recent DL based method DCP~\cite{wang2019deep} replaces the handcrafted feature descriptor with a CNN. DeepGMR~\cite{yuan2020deepgmr} further estimates the points-to-components correspondences in the latent GMM.
In summary, the main problem is that they require the inputs to have distinctive geometric structures, so as to promote sparse matched points. However, not all regions are distinctive, resulting in a limited number of matches or poor distributions. In addition, the transformation is calculated only from matched sparse points and their local neighbors, leaving the rest of the points untouched. In contrast, our work can use the predicted overlapping regions to aggregate global features.
\begin{figure*}[t]
    \centering
    \includegraphics[width=1.0\textwidth]{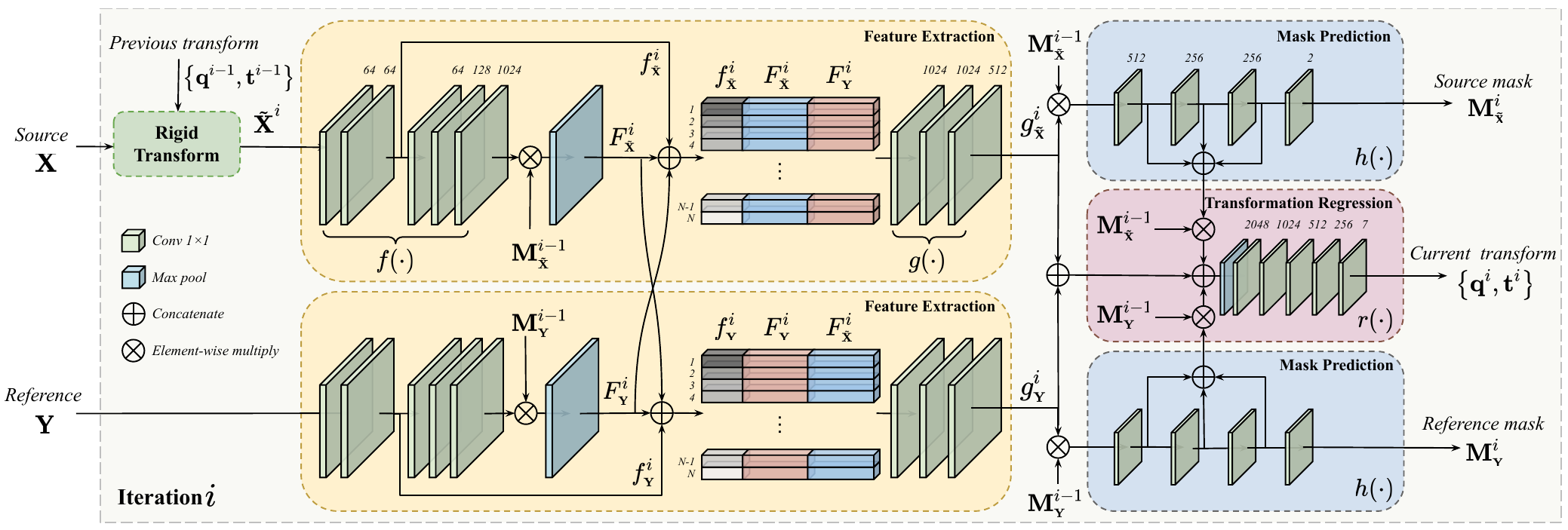}\\
    \caption{The overall architecture of our OMNet. During process of feature extraction, the global features $F_{\tilde{\mathbf{X}}}$ and $F_{\mathbf{Y}}$ are duplicated N times to concatenate with the point-wise features $f_{\tilde{\mathbf{X}}}$ and $f_{\mathbf{Y}}$, where N is the number of points in the inputs. The same background color denotes sharing weights. Superscripts denote the iteration count.}
    \vspace{-0.37cm}
    \label{fig:pipeline}
\end{figure*}
\vspace{-0.4cm}
\paragraph{Global Feature based Methods.}
Different from correspondence matching based methods, the previous global feature based methods compute rigid transformation from the entire point clouds (including overlapping and non-overlapping regions) of the two inputs without correspondences. PointNetLK~\cite{aoki2019pointnetlk} pioneers these methods, which adapts PointNet~\cite{qi2017pointnet} with the Lucas \&Kanade (LK) algorithm~\cite{lucas1981iterative} into a recurrent neural network. PCRNet~\cite{sarode2019pcrnet} improves the robustness against the noise by alternating the LK algorithm with a regression network. Furthermore, FMR~\cite{huang2020feature} adds a decoder branch and optimizes the global feature distance of the inputs. However, they all ignore the negative effect of the non-overlapping points and fail to register partial-to-partial inputs. Our network can deal with partiality and shows robustness to different partial manners.

\vspace{-0.35cm}
\paragraph{Partial-to-partial Registration Methods.}Registration of partial-to-partial point clouds is presented as a more realistic problem by recent works~\cite{wang2019prnet, yew2020-RPMNet, idam}. In particular, PRNet~\cite{wang2019prnet} extends DCP as an iterative pipeline and tackles the partiality by detecting keypoints. Moreover, the learnable Gumble-Softmax~\cite{jang2016categorical} is used to control the smoothness of the matching matrix. RPMNet~\cite{yew2020-RPMNet} further utilizes Sinkhorn normalization~\cite{sinkhorn1964relationship} to encourage the bijectivity of the matching matrix. However, they suffer from the same problem as the correspondence matching based methods, i.e., they can only use sparse points. In contrast, our method can utilize information from the entire overlapping points.

\section{Method} \label{sec:3}
Our pipeline is illustrated in Fig.~\ref{fig:pipeline}. We represent the 3D transformation in the form of quaternion $\mathbf{q}$ and translation $\mathbf{t}$.
At each iteration $i$, the source point cloud $\mathbf{X}$ is transformed by the rigid transformation $\left\{\mathbf{q}^{i-1}, \mathbf{t}^{i-1}\right\}$ estimated from the previous step into the transformed point cloud $\tilde{\mathbf{X}}^{i}$. Then, the global features of two point clouds are extracted by the feature extraction module (Sec.~\ref{sec:3.1}). Concurrently, the hybrid features from two point clouds are fused and fed to an overlapping mask prediction module (Sec.~\ref{sec:3.2}) to segment the overlapping region. Meanwhile, a transformation regression module (Sec.~\ref{sec:3.3}) takes the fused hybrid features as input and outputs the transformation $\left\{\mathbf{q}^{i}, \mathbf{t}^{i}\right\}$ for the next iteration. Finally, the loss functions are detailed in Sec.~\ref{sec:3.4}.
\subsection{Global Feature Extraction} \label{sec:3.1}
The feature extraction module aims to learn a function $f(\cdot)$, which can generate distinctive global features $F_{\scriptstyle \mathbf{X}}$ and $F_{\scriptstyle \mathbf{Y}}$ from the source point cloud $\mathbf{X}$ and the reference point cloud $\mathbf{Y}$ respectively. An important requirement is that the orientation and the spatial coordinates of the original input should be maintained, so that the rigid transformation can be estimated from the difference between the two global features. 
Inspired by PointNet~\cite{qi2017pointnet}, at each iteration, the global features of input $\tilde{\mathbf{X}}^{i}$ and $\mathbf{Y}$ are given by:
\begin{equation}
    \small
    F_{\beta}^{i}=\operatorname{max}\{\mathbf{M}^{i-1}_{\beta}\cdot{f\left(\beta\right)}\},\quad \beta \in\{\tilde{\mathbf{X}}^{i},\mathbf{Y}\},
\end{equation}
where $f(\cdot)$ denotes a multi-layer perceptron network (MLP), which is fed with $\tilde{\mathbf{X}}^{i}$ and $\mathbf{Y}$ to generate point-wise features $f^{i}_{\scriptstyle \tilde{\mathbf{X}}}$ and $f^{i}_{\scriptstyle \mathbf{Y}}$. $\mathbf{M}^{i-1}_{\scriptstyle \tilde{\mathbf{X}}}$ and $\mathbf{M}^{i-1}_{\scriptstyle \mathbf{Y}}$ are the overlapping masks of ${\tilde{\mathbf{X}}}^{i}$ and $\mathbf{Y}$, which are generated by the previous step and detailed in Sec.~\ref{sec:3.2}. The point-wise features $f_{\scriptstyle \tilde{\mathbf{X}}}$ and $f_{\scriptstyle \mathbf{Y}}$ are aggregated by a max-pool operation $\operatorname{max}\{\cdot\}$, which can deal with an arbitrary number of orderless points. 
\subsection{Overlapping Mask Prediction} \label{sec:3.2}
\begin{figure}[t]
    \centering
    \includegraphics[width=1.0\linewidth]{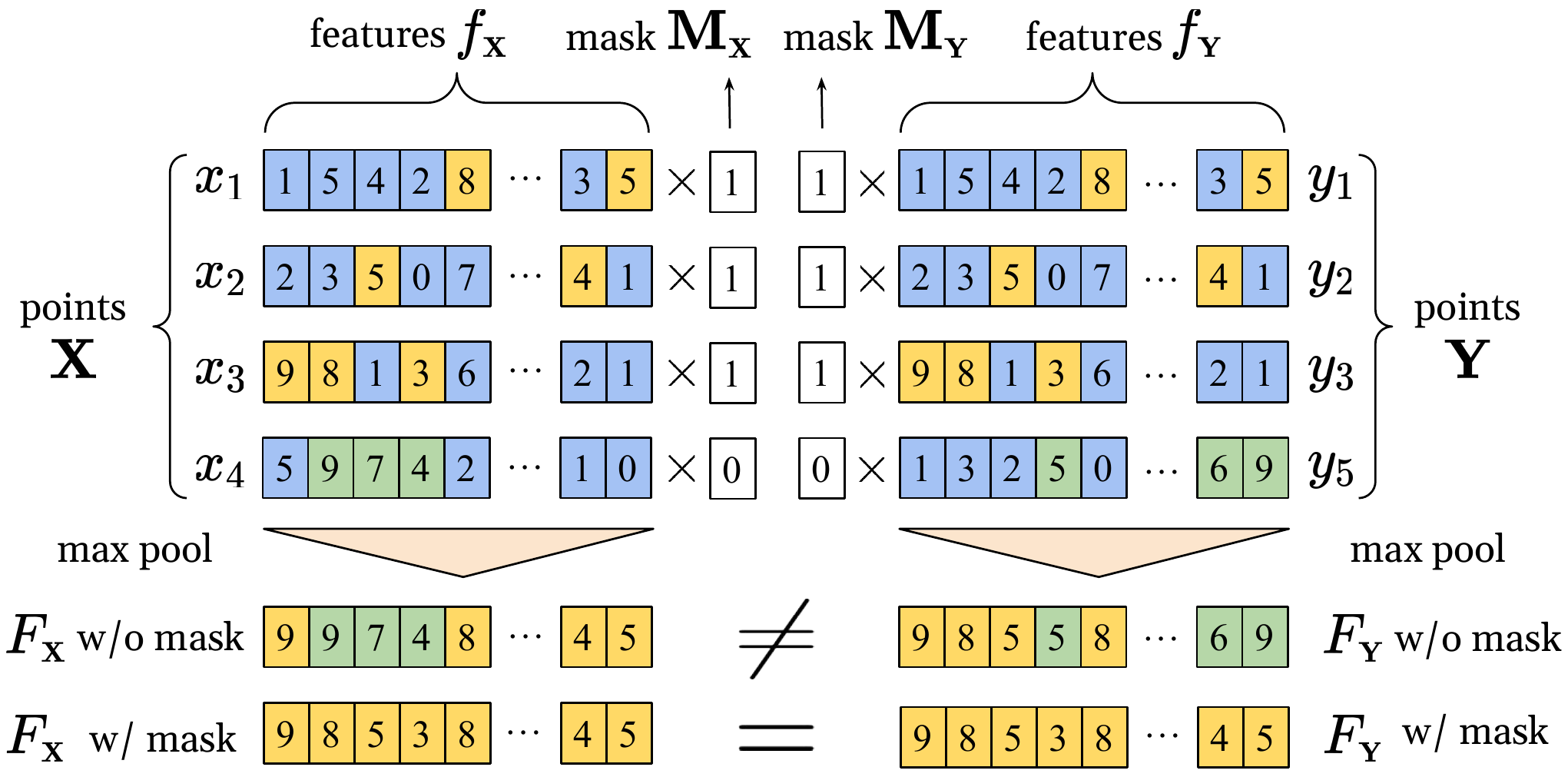}\\
    \caption{We show 4 orderless points of each point cloud. The same subscript denotes the corresponding points. Yellow indicates the maximum of each channel in the features of overlapping points and green indicates the interference of non-overlapping points. The global features of $\mathbf{X}$ and $\mathbf{Y}$ are the same only when they are weighted by the masks $\mathbf{M}_{\scriptstyle \mathbf{X}}$ and $\mathbf{M}_{\scriptstyle \mathbf{Y}}$.}
    \vspace{-0.1cm}
    \label{fig:mask}
\end{figure}

In partial-to-partial scenes, especially those including the noise, there exists non-overlapping regions between the input point clouds $\mathbf{X}$ and $\mathbf{Y}$. However, not only does it have no contributions to the registration procedure, but it also interferes to the global feature extraction, as shown in Fig.~\ref{fig:mask}. RANSAC~\cite{fischler1981random} is widely adopted in traditional methods to find the inliers when solving the most approximate matrix for the scene alignment. Following a similar idea, we propose a mask prediction module to segment the overlapping region automatically. Refer to PointNet~\cite{qi2017pointnet}, point segmentation only takes one point cloud as input and requires a combination of local and global knowledge. However, overlapping region prediction requires additional geometric information from both two input point clouds $\mathbf{X}$ and $\mathbf{Y}$. We can achieve this in a simple yet highly effective manner.

Specifically, at each iteration, the global features $F^{i}_{\scriptstyle \tilde{\mathbf{X}}}$ and $F^{i}_{\scriptstyle \mathbf{Y}}$ are fed back to point-wise features by concatenating with each of the point features $f^{i}_{\scriptstyle \tilde{\mathbf{X}}}$ and $f^{i}_{\scriptstyle \mathbf{Y}}$ accordingly. Then, a MLP $g(\cdot)$ is applied to fuse the above hybrid features, which are further used to segment overlapping regions and regress the rigid transformation. So we can obtain two overlapping masks $\mathbf{M}^{i}_{\scriptstyle \tilde{\mathbf{X}}}$ and $\mathbf{M}^{i}_{\scriptstyle \mathbf{Y}}$ as,
\begin{equation}
    \small
    \mathbf{M}^{i}_{\scriptstyle \tilde{\mathbf{X}}}=h\left(g\left({f^{i}_{\scriptstyle \tilde{\mathbf{X}}}}{\;\oplus\;}{F^{i}_{\scriptstyle \tilde{\mathbf{X}}}}{\;\oplus\;}{F^{i}_{\scriptstyle \mathbf{Y}}}\right)\cdot{\mathbf{M}^{i-1}_{\scriptstyle \tilde{\mathbf{X}}}}\right),
    \end{equation}
    \begin{equation}
    \small
    \mathbf{M}^{i}_{\scriptstyle \mathbf{Y}}=h\left(g\left({f^{i}_{\scriptstyle \mathbf{Y}}}{\;\oplus\;}{F^{i}_{\scriptstyle \mathbf{Y}}}{\;\oplus\;}{F^{i}_{\scriptstyle \tilde{\mathbf{X}}}}\right)\cdot{\mathbf{M}^{i-1}_{\scriptstyle \mathbf{Y}}}\right),
\end{equation}
where $h(\cdot)$ denotes the overlapping prediction network, which consists of several convolutional layers followed by a softmax layer. We define the fused point-wise features of the inputs $\mathbf{X}$ and $\mathbf{Y}$ produced by $g(\cdot)$ as $g_{\scriptstyle \mathbf{X}}$ and $g_{\scriptstyle \mathbf{Y}}$. $\oplus$ denotes the concatenation operation.
\subsection{Rigid Transformation Regression} \label{sec:3.3}
Given the point-wise features $g^{i}_{\scriptstyle \tilde{\mathbf{X}}}$ and $g^{i}_{\scriptstyle \mathbf{Y}}$ at each iteration $i$, we concatenate them with the features outputting from intermediate layers of the overlapping mask prediction module. Therefore, the features used to regress transformation can be enhanced by the classification information in the mask prediction branch. Meanwhile, the features used to predict the masks benefit from the geometric knowledge in the transformation branch. Then, the concatenated features are fed to the rigid transformation regression network, which produces a 7D vector, with the first 3 values of the 7D vector we use to represent the translation vector $\mathbf{t} \in \mathbb{R}^{3}$ and the last 4 values represent the 3D rotation in the form of quaternion~\cite{shoemake1985animating} $\mathbf{q} \in \mathbb{R}^{4}, \mathbf{q}^{T} \mathbf{q}=1$. $r(\cdot)$ represents the whole process in every iteration $i$, i.e.
\begin{equation}
    \small
    \hspace{-2.9mm}\left\{\mathbf{q}^{i},\mathbf{t}^{i}\right\}=r\!\left(\operatorname{max}\{{g^{i}_{\scriptstyle \tilde{\mathbf{X}}}}\oplus{h^{i}_{\scriptstyle \tilde{\mathbf{X}}}}\!\cdot\!{\mathbf{M}^{i-1}_{\scriptstyle \tilde{\mathbf{X}}}}\oplus{g^{i}_{\scriptstyle \mathbf{Y}}}\oplus{h^{i}_{\scriptstyle \mathbf{Y}}}\!\cdot\!{\mathbf{M}^{i-1}_{\scriptstyle \mathbf{Y}}}\}\right),
\end{equation}
where $h^{i}_{\scriptstyle \tilde{\mathbf{X}}}$ and $h^{i}_{\scriptstyle \mathbf{Y}}$ are the concatenated features from the mask prediction branch. $\mathbf{M}^{i-1}_{\scriptstyle \tilde{\mathbf{X}}}$ and $\mathbf{M}^{i-1}_{\scriptstyle \mathbf{Y}}$ are used to eliminate the interference of the non-overlapping points.

After $N$ iterations, we obtain the overall transformation between the two inputs by accumulating all the estimated transformations at each iteration.

\begin{figure}[t]
   \centering
   \includegraphics[width=1.0\linewidth]{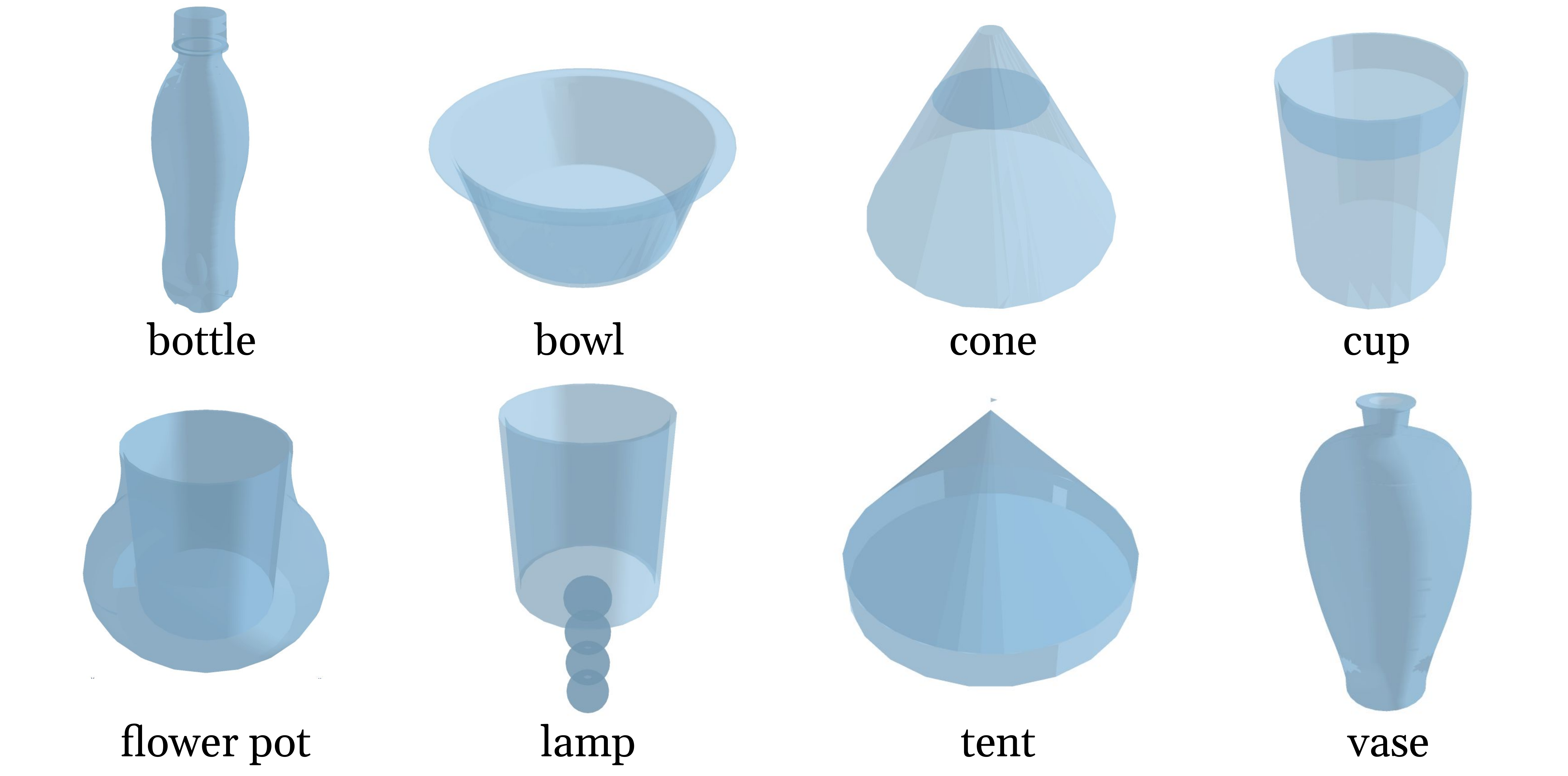}\\
   \caption{Example CAD models of 8 axisymmetrical categories.}
   \vspace{-0.1cm}
   \label{fig:axisymmetrical categories}
\end{figure}
\subsection{Loss Functions} \label{sec:3.4}
We simultaneously predict overlapping masks and estimate rigid transformations, so that two loss functions are proposed to supervise the above two procedures separately.

\vspace{-0.35cm}
\paragraph{Mask Prediction Loss.}
The goal of mask prediction loss is to segment the overlapping region in the input point clouds $\mathbf{X}$ and $\mathbf{Y}$. To balance the contributions of positive and negative samples, the frequency weighted softmax cross-entropy loss is exploited at each iteration $i$, i.e.
\begin{equation}
    \small
    \mathcal{L}_{mask}\!=\!-\!\alpha \mathbf{M}^{i}_{g} \log (\mathbf{M}^{i}_{p})\!-\!(1\!-\!\alpha)(1\!-\!\mathbf{M}^{i}_{g}) \log(1\!-\!\mathbf{M}^{i}_{p}),
\end{equation}
where $\mathbf{M}_p$ denotes the probability of points belonging to the overlapping region, and $\alpha$ is the overlapping ratio of the inputs. We define the assumed mask label $\mathbf{M}_{g}$ to represent the overlapping region of the two inputs, which is computed by setting fixed threshold (set to 0.1) for the closest point distances between the source that transformed by the ground-truth transformation and reference. Each element is
\begin{equation}
    \small
    M_{g}=\left\{\begin{array}{ll}
    1 &\text{if point } \mathbf{x}_{j} \text{ corresponds to } \mathbf{y}_{k} \\
    0 &\text{otherwise}
    \end{array}\right..
\end{equation}
The current mask is estimated based on the previous mask, so the label needs to be recalculated for each iteration. 
\begin{table*}[t]
    \small
    \centering
    \resizebox{\linewidth}{!}{
        \begin{tabular}{clcccccccccccc}
            \toprule
            & \multicolumn{1}{c}{} & \multicolumn{2}{c}{RMSE(\textbf{R})} & \multicolumn{2}{c}{MAE(\textbf{R})} & \multicolumn{2}{c}{RMSE(\textbf{t})} & \multicolumn{2}{c}{MAE(\textbf{t})} & \multicolumn{2}{c}{Error(\textbf{R})} & \multicolumn{2}{c}{Error(\textbf{t})} 
            \\ 
            \cmidrule(lr){3-4} \cmidrule(lr){5-6} \cmidrule(lr){7-8} \cmidrule(lr){9-10} \cmidrule(lr){11-12} \cmidrule(lr){13-14} 
            &\multicolumn{1}{c}{\multirow{-2}{*}{Method}}&\emph{OS} &\emph{TS} &\emph{OS} &\emph{TS} &\emph{OS} &\emph{TS} &\emph{OS} &\emph{TS} &\emph{OS} &\emph{TS} &\emph{OS} &\emph{TS}
            \\ 
            \midrule
            & ICP~\cite{besl1992method}            & 21.043 & 21.246 & 8.464 & 9.431 & 0.0913 & 0.0975 & 0.0467 & 0.0519 & 16.460 & 17.625 & 0.0921 & 0.1030 \\
            & Go-ICP~\cite{yang2013go}             & 13.458 & 11.296 & 3.176 & 3.480 & 0.0462 & 0.0571 & 0.0149 & 0.0206 & 6.163 & 7.138 & 0.0299 & 0.0407 \\
            & Symmetric ICP~\cite{rusinkiewicz2019symmetric} & 5.333 & 6.875 & 4.787 & 6.069 & 0.0572 & 0.0745 & 0.0517 & 0.0668 & 9.424 & 12.103 & 0.0992 & 0.1290 \\ 
            & FGR~\cite{zhou2016fast}              & 4.741 & 28.865 & 1.110 & 16.168 & 0.0269 & 0.1380 & \textcolor{blue}{0.0070} & 0.0774 & 2.152 & 30.192 & \textcolor{blue}{0.0136} & 0.1530 \\
            & PointNetLK~\cite{aoki2019pointnetlk} & 16.429 & 14.888 & 7.467 & 7.603 & 0.0832 & 0.0842 & 0.0443 & 0.0464 & 14.324 & 14.742 & 0.0880 & 0.0920 \\
            & DCP~\cite{wang2019deep}              & 4.291 & 5.786 & 3.006 & 3.872 & 0.0426 & 0.0602 & 0.0291 & 0.0388 & 5.871 & 7.903 & 0.0589 & 0.0794 \\
            & PRNet~\cite{wang2019prnet}           & \textcolor{blue}{1.588} & 3.677 & \textcolor{blue}{0.976} & 2.201 & \textcolor{blue}{0.0146} & 0.0307 & 0.0101 & 0.0204 & \textcolor{blue}{1.871} & 4.223 & 0.0201 & 0.0406 \\
            & FMR~\cite{huang2020feature}          & 2.740 & \textcolor{blue}{3.456} & 1.448 & \textcolor{blue}{1.736} & 0.0250 & \textcolor{blue}{0.0292} & 0.0112 & \textcolor{blue}{0.0138} & 2.793 & \textcolor{blue}{3.281} & 0.0218 & \textcolor{blue}{0.0272} \\
            & IDAM~\cite{idam}                     & 4.744 & 7.456 & 1.346 & 4.387 & 0.0395 & 0.0604 & 0.0108 & 0.0352 & 2.610 & 8.577 & 0.0216 & 0.0698 \\
            & DeepGMR~\cite{yuan2020deepgmr}       & 13.266 & 21.985 & 6.883 & 11.113 & 0.0748 & 0.0936 & 0.0476 & 0.0587 & 13.536 & 20.806 & 0.0937 & 0.1171 \\
            \multirow{-10}{*}{\rotatebox{90}{(a) Unseen Shapes}} & Ours & \textcolor{red}{0.898} & \textcolor{red}{1.045} & \textcolor{red}{0.325} & \textcolor{red}{0.507} & \textcolor{red}{0.0078} & \textcolor{red}{0.0084} & \textcolor{red}{0.0049} & \textcolor{red}{0.0056} & \textcolor{red}{0.639} & \textcolor{red}{0.991} & \textcolor{red}{0.0099} & \textcolor{red}{0.0112} 
            \\ 
            \midrule
            & ICP~\cite{besl1992method}            & 17.236 & 18.458 & 8.610 & 9.335 & 0.0817 & 0.0915 & 0.0434 & 0.0505 & 16.824 & 18.194 & 0.0855 & 0.0993 \\
            & Go-ICP~\cite{yang2013go}             & 13.572 & 14.162 & 3.416 & 4.190 & 0.0448 & 0.0533 & 0.0152 & 0.0206 & 6.688 & 8.286 & 0.0299 & 0.0409 \\
            & Symmetric ICP~\cite{rusinkiewicz2019symmetric} & 6.599 & 7.415 & 5.962 & 6.552 & 0.0654 & 0.0759 & 0.0592 & 0.0684 & 11.713 & 13.113 & 0.1134 & 0.1315 \\ 
            & FGR~\cite{zhou2016fast}              & 6.390 & 29.838 & \textcolor{blue}{1.240} & 16.361 & 0.0375 & 0.1470 & \textcolor{blue}{0.0081} & 0.0818 & \textcolor{blue}{2.204} & 31.153 & \textcolor{blue}{0.0156} & 0.1630 \\
            & PointNetLK~\cite{aoki2019pointnetlk} & 18.294 & 21.041 & 9.730 & 10.740 & 0.0917 & 0.1130 & 0.0526 & 0.0629 & 18.845 & 20.438 & 0.1042 & 0.1250 \\
            & DCP~\cite{wang2019deep}              & 6.754 & 7.683 & 4.366 & 4.747 & 0.0612 & 0.0675 & 0.0403 & 0.0427 & 8.566 & 9.764 & 0.0807 & 0.0862 \\
            & PRNet~\cite{wang2019prnet}           & \textcolor{blue}{2.712} & 6.506 & 1.372 & 3.472 & \textcolor{red}{0.0171} & 0.0388 & 0.0118 & 0.0257 & 2.607 & 6.789 & 0.0237 & 0.0510 \\
            & FMR~\cite{huang2020feature}          & 5.041 & \textcolor{blue}{5.119} & 2.304 & \textcolor{blue}{2.349} & 0.0383 & \textcolor{blue}{0.0296} & 0.0158 & \textcolor{blue}{0.0147} & 4.525 & \textcolor{blue}{4.553} & 0.0314 & \textcolor{blue}{0.0292} \\
            & IDAM~\cite{idam}                     & 6.852 & 8.346 & 1.761 & 4.540 & 0.0540 & 0.0590 & 0.0138 & 0.0329 & 3.433 & 8.679 & 0.0275 & 0.0656 \\
            & DeepGMR~\cite{yuan2020deepgmr}       & 18.890 & 23.472 & 9.322 & 12.863 & 0.0870 & 0.0987 & 0.0559 & 0.0658 & 17.513 & 24.425 & 0.1108 & 0.1298 \\
            \multirow{-10}{*}{\rotatebox{90}{(b) Unseen Categories}} & Ours & \textcolor{red}{2.079} & \textcolor{red}{2.514} & \textcolor{red}{0.619} & \textcolor{red}{1.004} & \textcolor{blue}{0.0177} & \textcolor{red}{0.0147} & \textcolor{red}{0.0077} & \textcolor{red}{0.0078} & \textcolor{red}{1.241} & \textcolor{red}{1.949} & \textcolor{red}{0.0154} & \textcolor{red}{0.0154} 
            \\ 
            \midrule
            & ICP~\cite{besl1992method}            & 19.945 & 21.265 & 8.546 & 9.918 & 0.0898 & 0.0966 & 0.0482 & 0.0541 & 16.599 & 18.540 & 0.0949 & 0.1070 \\
            & Go-ICP~\cite{yang2013go}             & 13.612 & 12.337 & 3.655 & 3.880 & 0.0489 & 0.0560 & 0.0174 & 0.0218 & 7.257 & 7.779 & 0.0348 & 0.0433 \\
            & Symmetric ICP~\cite{rusinkiewicz2019symmetric} & 5.208 & 6.769 & 4.703 & 5.991 & 0.0518 & 0.0680 & 0.0462 & 0.0609 & 9.174 & 11.895 & 0.0897 & 0.1178 \\ 
            & FGR~\cite{zhou2016fast}              & 22.347 & 34.035 & 10.309 & 19.188 & 0.1070 & 0.1601 & 0.0537 & 0.0942 & 19.934 & 35.775 & 0.1068 & 0.1850 \\
            & PointNetLK~\cite{aoki2019pointnetlk} & 20.131 & 22.399 & 11.864 & 13.716 & 0.0972 & 0.1092 & 0.0516 & 0.0601 & 18.552 & 20.250 & 0.1032 & 0.1291 \\
            & DCP~\cite{wang2019deep}              & 4.862 & 4.775 & 3.433 & 2.964 & 0.0486 & 0.0474 & 0.0340 & 0.0300 & 6.653 & 6.024 & 0.0690 & 0.0616 \\
            & PRNet~\cite{wang2019prnet}           & \textcolor{blue}{1.911} & \textcolor{blue}{3.197} & \textcolor{blue}{1.213} & \textcolor{blue}{2.047} & \textcolor{blue}{0.0180} & 0.0294 & \textcolor{blue}{0.0123} & 0.0195 & \textcolor{blue}{2.284}                & \textcolor{blue}{3.932} & \textcolor{blue}{0.0245} & 0.0392 \\
            & FMR~\cite{huang2020feature}          & 2.898 & 3.551 & 1.747 & 2.178 & 0.0246 & \textcolor{blue}{0.0273} & 0.0133 & \textcolor{blue}{0.0155} & 3.398 & 4.200 & 0.0260 & \textcolor{blue}{0.0307} \\
            & IDAM~\cite{idam}                     & 5.551 & 6.846 & 2.990 & 3.997 & 0.0486 & 0.0563 & 0.0241 & 0.0318 & 5.741 & 7.810 & 0.0480 & 0.0629 \\
            & DeepGMR~\cite{yuan2020deepgmr}       & 17.693 & 20.433 & 8.578 & 10.964 & 0.0849 & 0.0944 & 0.0531 & 0.0593 & 16.504 & 20.830 & 0.1048 & 0.1183 \\
            \multirow{-10}{*}{\rotatebox{90}{\makecell[c]{(c) Gaussian Noise}}} & Ours & \textcolor{red}{1.009} & \textcolor{red}{1.305} & \textcolor{red}{0.548} & \textcolor{red}{0.757} & \textcolor{red}{0.0089} & \textcolor{red}{0.0103} & \textcolor{red}{0.0061} & \textcolor{red}{0.0075} & \textcolor{red}{1.076} & \textcolor{red}{1.490} & \textcolor{red}{0.0123} & \textcolor{red}{0.0149} 
            \\ 
            \bottomrule
        \end{tabular}%
    }
    \caption{Results on ModelNet40. For each metric, the left column \emph{OS} denotes the results on the original once-sampled data, and the right column \emph{TS} denotes the results on our twice-sampled data. Red indicates the best performance and blue indicates the second-best result.}
    \vspace{-0.37cm}
    \label{tab:1-3}
\end{table*}

\vspace{-0.35cm}
\paragraph{Transformation Regression Loss.}
Benefiting from the continuity of the quaternions, it is able to employ a fairly straightforward strategy for training, measuring the deviation of $\left\{\mathbf{q}, \mathbf{t}\right\}$ from ground truth for the generated point cloud pairs. So the transformation regression loss at iteration $i$ is
\begin{equation}
    \small
    \mathcal{L}_{reg}=\abs{\mathbf{q}^{i}-\mathbf{q}_{g}}+\lambda \norm{\mathbf{t}^{i}-\mathbf{t}_{g}},
\end{equation}
where subscript $g$ denotes ground-truth. We notice that using the combination of $\ell^{1}$ and $\ell^{2}$ distance can marginally improve performance during the training and the inference. $\lambda$ is empirically set to 4.0 in most of our experiments.

The overall loss is the sum of the two losses:  
\vspace{-0.1cm}
\begin{equation}
    \small
    \mathcal{L}_{total}=\mathcal{L}_{mask}+ \mathcal{L}_{reg}.
    \vspace{-0.1cm}
\end{equation}

We compute the loss for every iteration, and they have equal contribution to the final loss during training.

\section{Experiments} \label{sec:4}
In this section, we first describe the pre-processing for the datasets and the implementation details of our method in Sec.~\ref{sec:4.1}. Concurrently, the experimental settings of competitors are presented in Sec.~\ref{sec:4.2}. Moreover, we show the results for different experiments to demonstrate the effectiveness and robustness of our method in Sec.~\ref{sec:4.3} and Sec.~\ref{sec:4.4}. Finally, we perform ablation studies in Sec.~\ref{sec:4.5}. 
\subsection{Dataset and Implementation Details} \label{sec:4.1}
\paragraph{ModelNet40.}
We use the ModelNet40 dataset to test the effectiveness following~\cite{aoki2019pointnetlk, wang2019deep, sarode2019pcrnet, wang2019prnet, huang2020feature, yew2020-RPMNet, idam}. The ModelNet40 contains CAD models from 40 categories. Previous works use processed data from PointNet~\cite{qi2017pointnet}, which has two issues when adopted to registration task: (1) for each object, it only contains 2,048 points sampled from the CAD model. However, in realistic scenes, the points in $\mathbf{X}$ have no exact correspondences in $\mathbf{Y}$. Training on this data cause over-fitting issue even adding noise or resampling, which is demonstrated by the experiment shown in our supplementary; (2) it involves some axisymmetrical categories, including \emph{bottle}, \emph{bowl}, \emph{cone}, \emph{cup}, \emph{flower pot}, \emph{lamp}, \emph{tent} and \emph{vase}, Fig.~\ref{fig:axisymmetrical categories} shows some examples. However, giving fixed ground-truths to axisymmetrical data is illogical, since it is possible to obtain arbitrary angles on the symmetrical axis for accurate registration. Fixing the label on symmetrical axis makes no sense. 
In this paper, we propose a proper manner to generate data. Specifically, we uniformly sample 2,048 points from each CAD model 40 times with different random seeds, then randomly choose 2 of them as $\mathbf{X}$ and $\mathbf{Y}$. It guarantees that we can obtain $C_{40}^{2}=780$ combinations for each object. We denote the data that points are sampled only once from CAD models as \textbf{once-sampled (\emph{OS})} data, and refer our data as \textbf{twice-sampled (\emph{TS})} data. Moreover, we simply remove the axisymmetrical categories.

To evaluate the effectiveness and robustness of our network, we use the official train and test splits of the first 14 categories (\emph{bottle}, \emph{bowl}, \emph{cone}, \emph{cup}, \emph{flower pot} and \emph{lamp} are removed) for training and validation respectively, and the test split of the remaining 18 categories (\emph{tent} and \emph{vase} are removed) for test. This results in 4,196 training, 1,002 validation, and 1,146 test models. Following previous works~\cite{wang2019deep, wang2019prnet, huang2020feature, yew2020-RPMNet, idam}, we randomly generate three Euler angle rotations within $\left[0^{\circ},45^{\circ}\right]$ and translations within $\left[-0.5,0.5\right]$ on each axis as the rigid transformation.

\vspace{-0.50cm}
\paragraph{Stanford 3D Scan.}
We use the Stanford 3D Scan dataset~\cite{curless1996volumetric} to test the generalizability of our method. The dataset has 10 real scans. The partial manner in PRNet~\cite{wang2019prnet} is applied to generate partially overlapping point clouds.

\vspace{-0.50cm}
\paragraph{7Scenes.}
7Scenes~\cite{shotton2013scene} is a widely used registration benchmark where data is captured by a Kinect camera in indoor environments. Following \cite{zeng20173dmatch, huang2020feature}, multiple depth images are projected into point clouds, then fused through truncated signed distance function (TSDF). The dataset is divided into 293 and 60 scans for training and test. The partial manner in PRNet~\cite{wang2019prnet} is applied.

\vspace{-0.50cm}
\paragraph{Implementation Details.}
Our network architecture is illustrated in Fig.~\ref{fig:pipeline}. We run $N=4$ iterations during training and test. Nevertheless, the $\{\mathbf{q},\mathbf{t}\}$ gradients are stopped at the beginning of each iteration to stabilize the training. Since the masks predicted by the first iteration may be inaccurate at the beginning of training, some overlapping points may be misclassified and affect the sequent iterations, we apply the masks after the second iteration. We train our network with Adam~\cite{kingma2014adam} optimizer for 260k iterations. The initial learning rate is 0.0001 and is multiplied by 0.1 after 220k iterations. The batch size is set to 64. 

\subsection{Baseline Algorithms} \label{sec:4.2}
We compare our method to traditional methods: ICP~\cite{besl1992method}, Go-ICP~\cite{yang2013go}, Symmetric ICP~\cite{rusinkiewicz2019symmetric}, FGR~\cite{zhou2016fast}, as well as recent DL based works: PointNetLK~\cite{aoki2019pointnetlk}, DCP~\cite{wang2019deep}, RPMNet~\cite{yew2020-RPMNet}, FMR~\cite{huang2020feature}, PRNet~\cite{wang2019prnet}, IDAM~\cite{idam} and DeepGMR~\cite{yuan2020deepgmr}. We use implementations of ICP and FGR in Intel Open3D~\cite{zhou2018open3d}, Symmetric ICP in PCL~\cite{Rusu_ICRA2011_PCL} and the others released by their authors. Moreover, the test set is fixed by setting random seeds. Note that the normals used in FGR and RPMNet are calculated after data pre-processing, which is slightly different from the implementation in RPMNet. We use the supervised version of FMR.

Following~\cite{wang2019deep, yew2020-RPMNet}, we measure anisotropic errors: root mean squared error (RMSE) and mean absolute error (MAE) of rotation and translation, and isotropic errors:
\begin{equation}
    \operatorname{Error}(\mathbf{R})=\angle\left(\mathbf{R}_{g}^{-1} \mathbf{R}_{p}\right),\;\operatorname{Error}(\mathbf{t})=\norm{\mathbf{t}_{g}-\mathbf{t}_{p}},
\end{equation}
where $\mathbf{R}_{g}$ and $\mathbf{R}_{p}$ denote the ground-truth and predicted rotation matrices converted from the quaternions $\mathbf{q}_{g}$ and $\mathbf{q}_{p}$ respectively. All  metrics should be zero if the rigid alignment is perfect. The angular metrics are in units of degrees.

\subsection{Evaluation on ModelNet40} \label{sec:4.3}
To evaluate the effectiveness of different methods, we conduct several experiments in this section. The data pre-processing settings of the first 3 experiments are the same as PRNet~\cite{wang2019prnet} and IDAM~\cite{idam}. In addition, the last experiment shows the robustness of our method to different partial manners, which is used in RPMNet~\cite{yew2020-RPMNet}.
\begin{figure*}[t]
    \centering
    \includegraphics[width=1.0\textwidth]{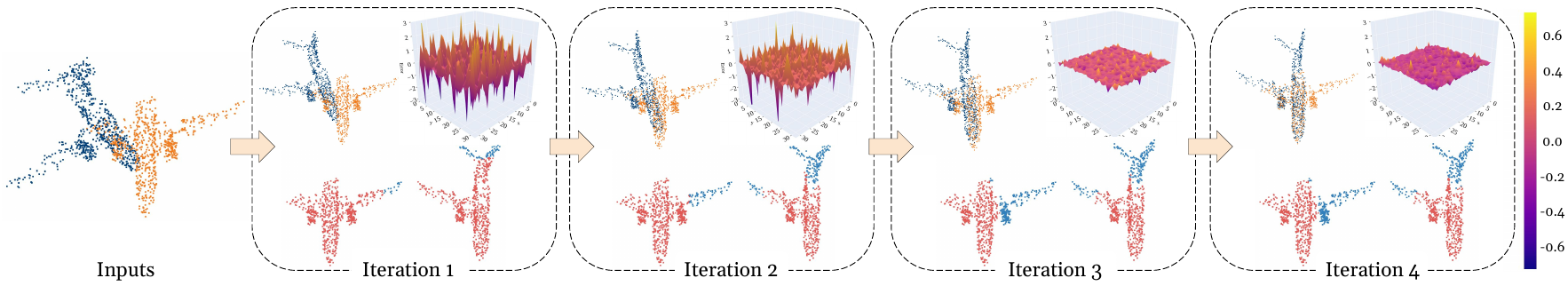}\\
    \caption{We show the registration result (top left), the difference between the global features of the inputs $\mathbf{X}$ and $\mathbf{Y}$ (top right), and the predicted masks (bottom) at each iteration. Red and blue indicate the predicted overlapping and non-overlapping regions respectively.}
    \vspace{-0.35cm}
    \label{fig:effect of mask}
\end{figure*}
\vspace{-0.50cm}
\paragraph{Unseen Shapes.}
In this experiment, we train models on training set of the first 14 categories and evaluate on validation set of the same categories without noise. Note that all points in $\mathbf{X}$ have \textbf{exact correspondences} in $\mathbf{Y}$ for the \emph{OS} data. We partial $\mathbf{X}$ and $\mathbf{Y}$ by randomly placing a point in space and computing its 768 nearest neighbors respectively, which is the same as used in~\cite{wang2019prnet, idam}.  All DL based methods are trained independently on both \emph{OS} and \emph{TS} data. Table~\ref{tab:1-3}(a) shows the results. 

\begin{figure*}[t]
    \centering
    \includegraphics[width=\linewidth]{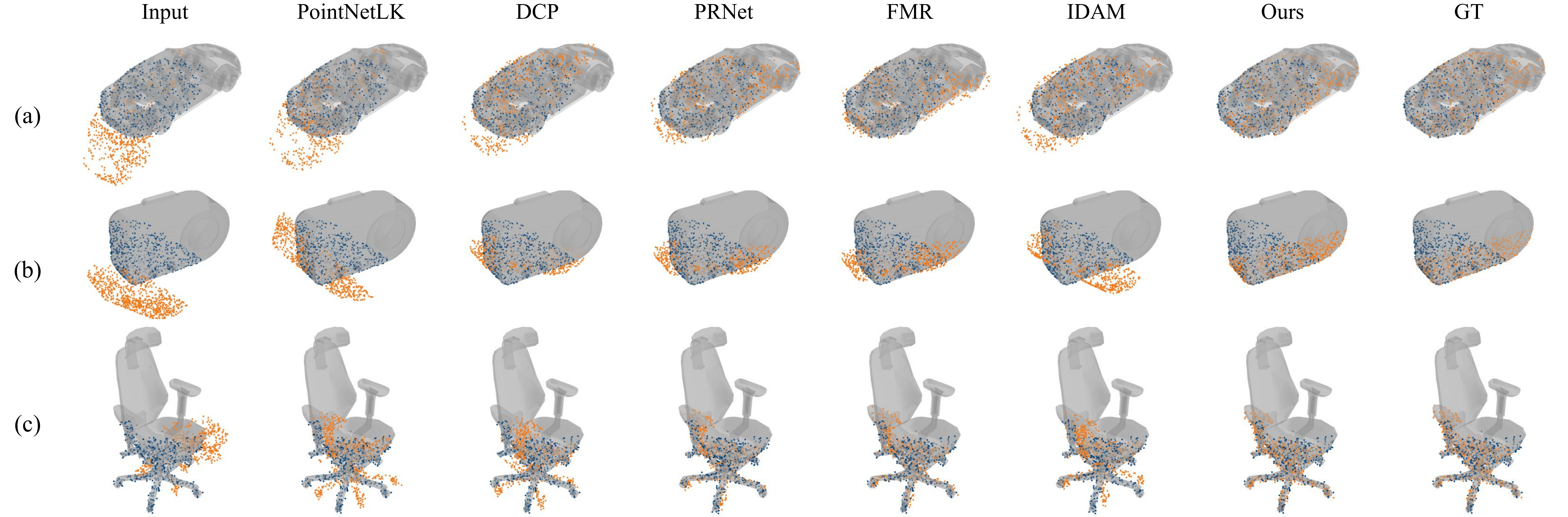}\\
    \caption{Example results on ModelNet40. (a) Unseen shapes, (b) Unseen categories, and (c) Unseen shapes with Gaussian noise.}
    \vspace{-0.37cm}
    \label{fig:qualitative results}
\end{figure*}

We can find that ICP~\cite{besl1992method} performs poorly because of the large difference in initial positions. Go-ICP~\cite{yang2013go} and FGR~\cite{zhou2016fast} achieve better performances, which are comparable to some DL based methods~\cite{aoki2019pointnetlk, wang2019deep, huang2020feature, idam}. Note that the large performance gap of FGR on two different data is caused by the calculation manner of normals. We use normals that are computed after data pre-processing, so that normals of $\mathbf{X}$ and $\mathbf{Y}$ are different in our \emph{TS} data. In addition, the results of IDAM~\cite{idam} are marginally worse than PRNet~\cite{wang2019prnet} because of the fixing manner of the test data, which is used in other DL based methods. Our method achieves very accurate registration and ranks first in all metrics. Example results on \emph{TS} data are shown in Fig.~\ref{fig:qualitative results}(a).

\vspace{-0.47cm}
\paragraph{Unseen Categories.}
We evaluate the performance on unseen categories without noise in this experiment. Models are trained on the first 14 categories and tested on the other 18 categories. The data pre-processing is the same as the first experiment. The results are summarized in Table~\ref{tab:1-3}(b). We can find that the performances of all DL based methods become marginally worse without training on the same categories. Nevertheless, traditional algorithms are not affected so much because of the handcrafted features. Our approach outperforms all the other methods. A qualitative comparison of the registration results can be found in Fig.~\ref{fig:qualitative results}(b).

\vspace{-0.47cm}
\paragraph{Gaussian Noise.}
In this experiment, we add noises that sampled from $\mathcal{N}(0,0.01^{2})$ and clipped to $\left[-0.05,0.05\right]$, then repeat the first experiment (unseen shapes). Table~\ref{tab:1-3}(c) shows the results. FGR is sensitive to noise, so that it performs much worse than the noise-free case. All DL based methods get worse with noises injected on the \emph{OS} data. The performances of correspondences matching based methods (DCP, PRNet and IDAM) show an opposite tendency on the \emph{TS} data compared to the global feature based methods (PointNetLK, FMR and ours), since the robustness of local feature descriptor is improved by the noise augmentation during training. Our method achieves the best performance. Example results are shown in Fig.~\ref{fig:qualitative results}(c).

\begin{table}[]
    \small
    \centering
    \setlength\tabcolsep{3pt} 
    \resizebox{\linewidth}{!}{%
        \begin{tabular}{lcccccc}
            \toprule
            Method     &RMSE(\textbf{R})       &MAE(\textbf{R})  &RMSE(\textbf{t})  &MAE(\textbf{t})  &Error(\textbf{R})      &Error(\textbf{t})       
            \\ 
            \midrule
            ICP~\cite{besl1992method} & 21.893 & 13.402 & 0.1963 & 0.1278 & 26.632 & 0.2679 \\
            Symmetric ICP~\cite{rusinkiewicz2019symmetric} & 12.576 & 10.987 & 0.1478 & 0.1203 & 21.807 & 0.2560 \\
            FGR~\cite{zhou2016fast} & 46.213 & 30.116 & 0.3034 & 0.2141 & 58.968 & 0.4364 \\
            PointNetLK~\cite{aoki2019pointnetlk} & 29.733 & 21.154 & 0.2670 & 0.1937 & 42.027 & 0.3964 \\
            DCP~\cite{wang2019deep} & 12.730 & 9.556 & 0.1072 & 0.0774 & 12.173 & 0.1586 \\
            RPMNet~\cite{yew2020-RPMNet} & \textcolor{blue}{6.160} & \textcolor{blue}{2.467} & \textcolor{blue}{0.0618} & \textcolor{blue}{0.0274} & \textcolor{blue}{4.913} & \textcolor{blue}{0.0589} \\
            FMR~\cite{huang2020feature} & 11.674 & 7.400 & 0.1364 & 0.0867 & 14.121 & 0.1870 \\ 
            Ours & \textcolor{red}{4.356} & \textcolor{red}{1.924} & \textcolor{red}{0.0486} & \textcolor{red}{0.0223} & \textcolor{red}{3.834} & \textcolor{red}{0.0476} 
            \\ 
            \bottomrule
        \end{tabular}%
    }
    \caption{Results on the twice-sampled (\emph{TS}) unseen categories with Gaussian noise using the partial manner of RPMNet.}
    \vspace{-0.37cm}
    \label{tab:different partial manners}
\end{table}

\vspace{-0.45cm}
\paragraph{Different Partial Manners.}
We notice that the previous works~\cite{wang2019prnet, yew2020-RPMNet} use different partial manners. To evaluate the effectiveness on different partial data, we also test the performance of different algorithms on the test set used in ~\cite{yew2020-RPMNet}. We retrain all DL based methods and show the results of the most difficult situation (unseen categories with Gaussian noise) in Table~\ref{tab:different partial manners}. For details about the partial manners, please refer to our supplementary.

\subsection{Evaluation on Real Data} \label{sec:4.4}
\begin{figure}[t]
    \centering
    \includegraphics[width=\linewidth]{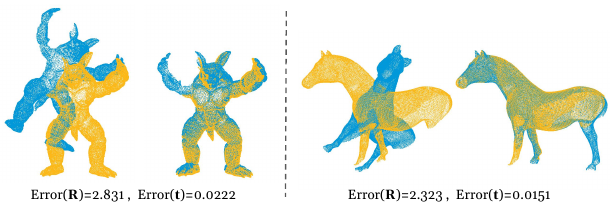}\\
    \caption{Example results on Stanford 3D Scan.}
    \vspace{-0.35cm}
    \label{fig:stanford 3d scan}
\end{figure}

\begin{figure}[t]
    \centering
    \includegraphics[width=\linewidth]{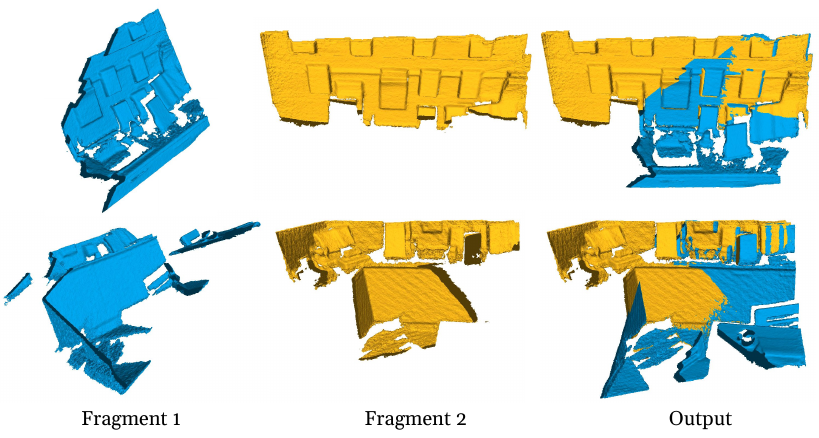}\\
    \caption{Example results on 7Scenes.}
    \vspace{-0.40cm}
    \label{fig:7scenes}
\end{figure}

To further evaluate the generalizability, we conduct experiments on the Stanford 3D Scan and 7Scenes datasets. Since the Stanford 3D Scan dataset only has 10 real scans, we directly use the model trained on the ModelNet40 without fine-tuning. Some qualitative examples are shown in Fig.~\ref{fig:stanford 3d scan}. Furthermore, we evaluate our method on the 7Scenes indoor dataset. The point clouds are normalized into the unit sphere. Our model is trained on 6 categories (\emph{Chess}, \emph{Fires}, \emph{Heads}, \emph{Office}, \emph{Pumpkin} and \emph{Stairs}) and tested on the other category (\emph{Redkitchen}). Fig.~\ref{fig:7scenes} shows some examples. For more results, please refer to our supplementary.

\begin{figure}[t]
    \centering
    \includegraphics[width=\linewidth]{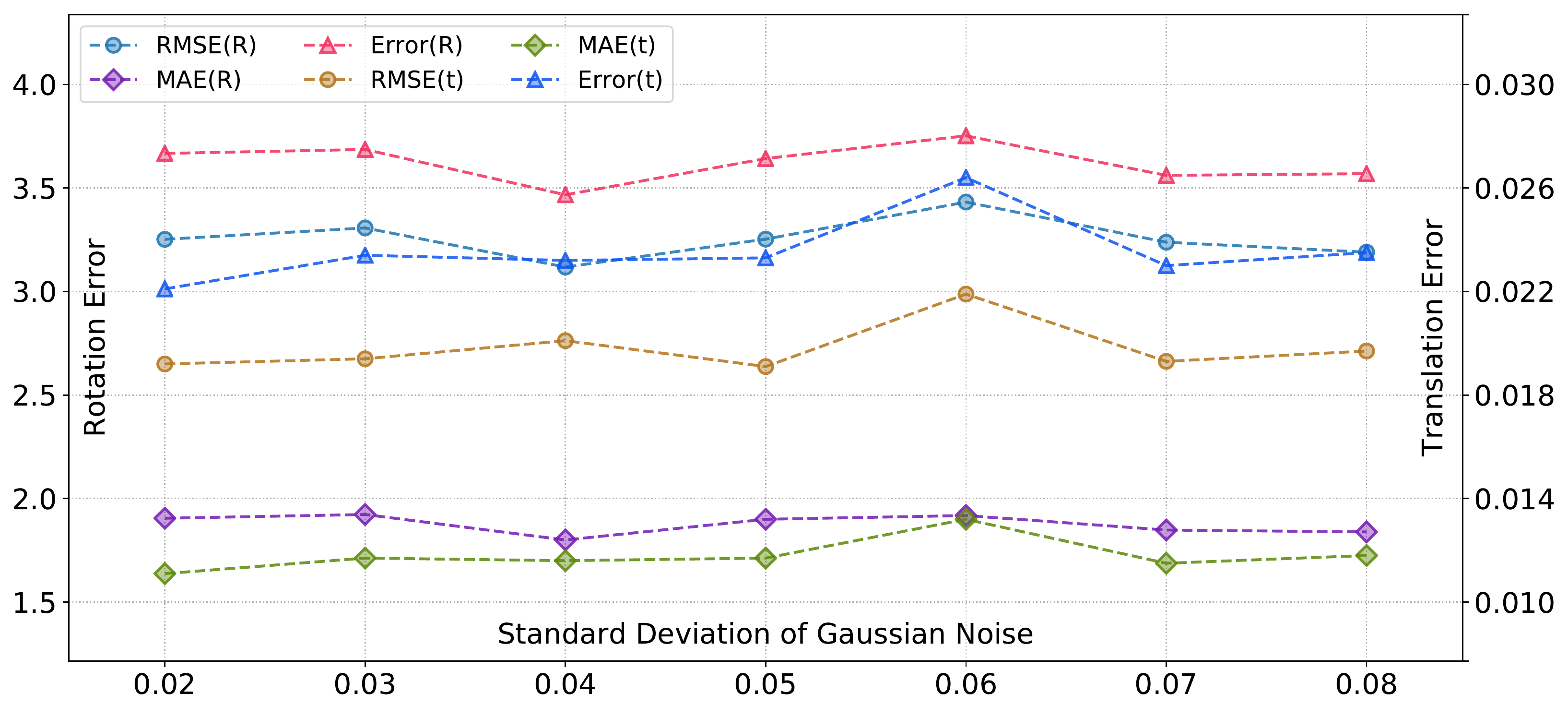}\\
    \caption{Errors of our method under different noise levels.}
    \vspace{-0.3cm}
    \label{fig:robustness against noise}
\end{figure}
\subsection{Ablation Studies}
\label{sec:4.5}
We perform ablation studies on the unseen shapes with noise \emph{TS} data to show the effectiveness of our components and settings. As shown in Table~\ref{tab:ablation_studies}, we denote our model that removed the following components as \textbf{B}aseline (B): \textbf{M}ask prediction module (M), \textbf{M}ask prediction \textbf{L}oss (ML), \textbf{F}usion layers (F) before the regression module, and \textbf{C}onnection (C) between mask prediction and regression modules. Besides, we only use top-k points based on the mask prediction probabilities to estimate rigid transformations. Moreover, we set different $\lambda$ in the loss function.
\begin{table}[t]
    \resizebox{\linewidth}{!}{%
    \begin{tabular}{lcccccccccccc}
        \toprule
        Model           & RMSE(\textbf{R}) & MAE(\textbf{R}) & RMSE(\textbf{t}) & MAE(\textbf{t}) & Error(\textbf{R}) & Error(\textbf{t}) \\
        \midrule
        B        & 3.216          & 2.751          & 0.0267          & 0.0232          & 5.250          & 0.0463          \\
        B+M           & 3.437          & 2.943          & 0.0349          & 0.0301          & 5.550          & 0.0604          \\
        B+M+ML      & 1.655          & 1.417          & 0.0158          & 0.0138          & 2.681          & 0.0274          \\
        B+M+ML+F         & 1.453          & 0.892          & 0.0111          & 0.0087          & 1.722          & 0.0171          \\
        B+M+ML+F+C     & \textbf{1.305} & \textbf{0.757} & \textbf{0.0103} & \textbf{0.0075} & \textbf{1.490} & \textbf{0.0149} \\
        \midrule
        Top-k, k=500    & 1.364          & 1.168          & 0.0127          & 0.0109          & 2.255          & 0.0220          \\
        Top-k, k=300    & 1.399          & 1.203          & 0.0161          & 0.0141          & 2.282          & 0.0278          \\
        Top-k, k=100    & 1.483          & 1.270          & 0.0180          & 0.0157          & 2.458          & 0.0311          \\
        \midrule
        $\lambda$=2.0   & 1.356          & 0.900          & 0.0109          & 0.0077          & 1.721          & 0.0154          \\
        $\lambda$=0.5   & 1.397          & 0.986          & 0.0116          & 0.0085          & 1.890          & 0.0169          \\
        $\lambda$=0.1   & 1.416          & 1.068          & 0.0127          & 0.0095          & 2.038          & 0.0189          \\
         \bottomrule
    \end{tabular}%
    }
    \caption{Ablation studies of each component and different settings.}
    \vspace{-0.37cm}
	\label{tab:ablation_studies}
\end{table}

We can see that without being supervised by the mask prediction loss, it has no improvement based on the baseline, which shows that the mask prediction can not be trained unsupervised. Comparing the third to the fifth lines with the baseline, we can find that all the components improve the performance. Since we do not estimate the matching candidates among the overlapping points, the top-k points from the source and reference may not be correspondent and distributed in the point clouds centrally, so that the results of top-k models are worse than using the entire masks. Furthermore, we adjust the $\lambda$ in the loss function. Since the data generation manner of~\cite{wang2019prnet, yew2020-RPMNet} constrain the translation within $[-0.5,0.5]$ as we use the $\ell^{2}$ loss for the translation, the translation loss is smaller than the quaternion, so that a large $\lambda$ aims to form comparable terms.

\section{Discussion}
In this section, we conduct several experiments to better understand how various settings affect our algorithm. 

\subsection{Effects of Mask}
To have a better intuition about the overlapping masks, we visualize the intermediate results in Fig.~\ref{fig:effect of mask}. We reshape the global feature vector of length 1,024 into a 32$\times$32 square matrix and compute the error between the transformed source $\tilde{\mathbf{X}}$ and reference $\mathbf{Y}$. At the first few iterations, the global feature differences are large, and the inputs are poorly aligned given inaccurate overlapping masks. With continuous iterating, the global feature difference becomes extremely small, while the alignment and predicted overlapping masks are almost perfect.

\begin{figure}[t]
    \centering
    \includegraphics[width=\linewidth]{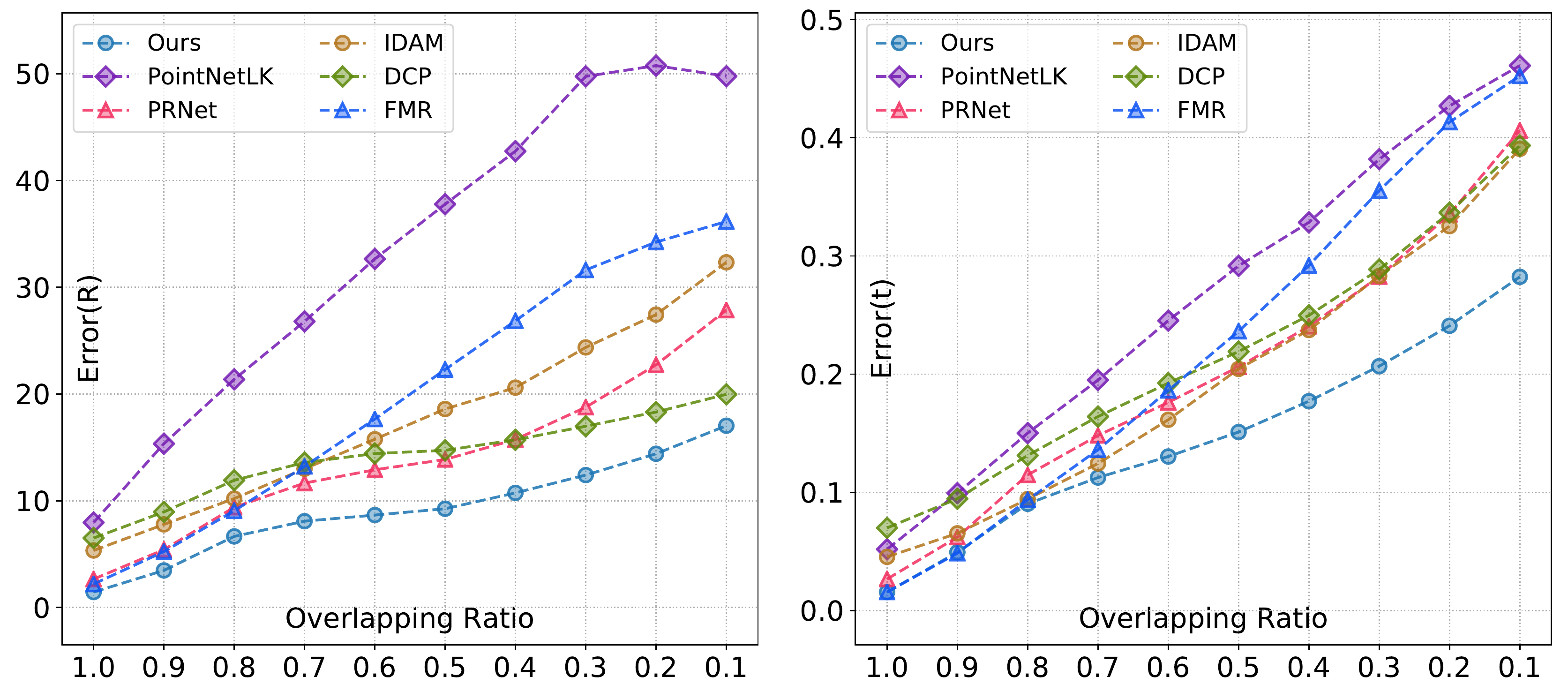}\\
    \caption{Isotropic errors of different overlapping ratios.}
    \vspace{-0.37cm}
    \label{fig:different overlapping ratio}
\end{figure}
\subsection{Robustness Against Noise}
To further demonstrate the robustness of our method, we train and test our models on ModelNet40 under different noise levels, as shown in Fig.~\ref{fig:robustness against noise}. We add noise sampled from $N(0, \sigma^{2})$ and clipped to $[-0.05, 0.05]$. The data is the same as the third experiment in Sec~\ref{sec:4.3}. Our method achieves comparable performance under various noise levels.

\subsection{Different Overlapping Ratio}
We test the best models of all methods from the first experiment in Sec.~\ref{sec:4.3} on the ModelNet40 \emph{TS} validation set with the overlapping ratio decreasing from 1.0 to 0.1. We first partial $\mathbf{X}$, then randomly select two adjacent parts from overlapping and non-overlapping regions of $\mathbf{Y}$. Fig.~\ref{fig:different overlapping ratio} shows the results. Unfortunately, DeepGMR fails to obtain sensible results. Our method shows the best performance.


\section{Conclusion}
We have presented the OMNet, a novel method for partial-to-partial point cloud registration. Previous global feature based methods pay less attention to partiality. They treat the input points equally, which are easily disturbed by the non-overlapping regions. Our method learns the overlapping masks to reject non-overlapping points for robust registration. Besides, we propose a practical data generation manner to solve the over-fitting issue and remove the axisymmetrical categories in the ModelNet40 dataset. Experiments show the state-of-the-art performance of our method.
\vspace{-0.9cm}
\paragraph{Acknowledgements.} 
This work is supported by the National Key R\&D Plan of the Ministry of Science and Technology (Project No.2020AAA0104400) and the National Natural Science Foundation of China (NSFC) under Grants No.62071097, No.61872067, No.62031009 and No.61720106004.

\clearpage
\appendix

\section{Overview}
This supplementary material provides more details on experiments in the main paper and includes more experiments to validate and analyze our proposed method.

In Sec.~\ref{sec:2}, we describe details in two data generation manners for point cloud registration, which are proposed by PRNet~\cite{wang2019prnet} and RPMNet~\cite{yew2020-RPMNet} separately. In Sec.~\ref{sec:3}, we show more experimental results including the performance on the validation and test sets, which are generated by the above two pre-processing manners.

\section{Details in Experiments}
\label{sec:2}
In this section, we describe two data preparation manners for the partial-to-partial point cloud registration. One of the  manners proposed by PRNet~\cite{wang2019prnet} is detailed in Sec.~\ref{sec:2.1}, while the other  used in RPMNet~\cite{yew2020-RPMNet} is described in Sec.~\ref{sec:2.2}. Fig.~\ref{fig:partial-to-partial data} illustrates some examples of the partial-to-partial data, which shows the difference between these two pre-processing manners.

\subsection{Data Generation Manner of PRNet}
\label{sec:2.1}
\begin{figure}[t]
\centering
\includegraphics[width=\linewidth]{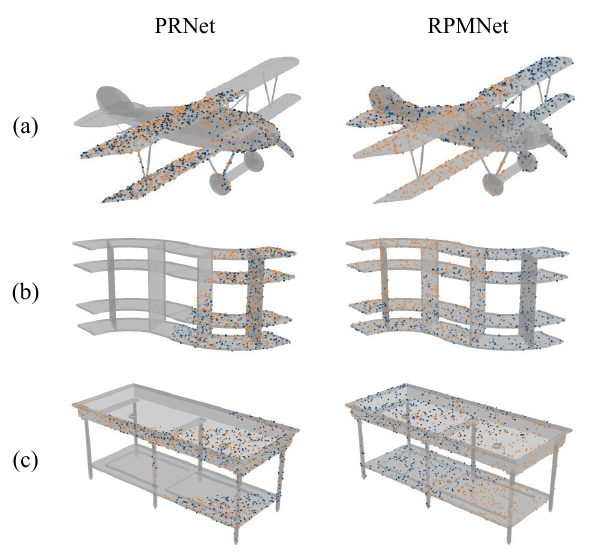}\\
\caption{Some examples of the partial-to-partial data. The left and right columns denote the point clouds that are processed in the manner of PRNet and RPMNet respectively. All of them are registered by the ground truth transformations. Blue denotes the source point cloud $\mathbf{X}$ and red denotes the reference point cloud $\mathbf{Y}$. The pairs of RPMNet are more decentralized than those of PRNet.}
\vspace{-0.37cm}
\label{fig:partial-to-partial data}
\end{figure}

We use this manner to generate data for the first three experiments on the ModelNet40 dataset in our main paper. Two point clouds are randomly chosen from 40 sampled point clouds as the source point cloud $\mathbf{X}$ and reference point cloud $\mathbf{Y}$ respectively, each of which contains 2,048 points. Along each axis, we randomly draw a rigid transformation: the rotation along each axis is sampled in $\left[0^{\circ},45^{\circ}\right]$ and the translation is in $\left[-0.5,0.5\right]$. The rigid transformation is applied to $\mathbf{Y}$, leading to $\mathbf{X}$. After that, we simultaneously partial $\mathbf{X}$ and $\mathbf{Y}$ by randomly placing a point in space and computing its 768 nearest neighbors in $\mathbf{X}$ and $\mathbf{Y}$ respectively. The left column in Fig.~\ref{fig:partial-to-partial data} shows some examples. However, the point clouds $\mathbf{X}$ and $\mathbf{Y}$ are similar in most cases, which means that the overlapping ratio is large.
\subsection{Data Generation Manner of RPMNet}
\label{sec:2.2}
We use this manner to generate data for the last experiment on the ModelNet40 dataset in our main paper. We use the same approach as which in RPMNet to obtain the source point cloud $\mathbf{X}$ and the reference point cloud $\mathbf{Y}$. For each point cloud, we sample a half-space with a random direction and shift it such that approximately 70\% of the points are retained. Then, the point clouds are downsampled to 717 points to maintain a similar point density as the previous experiments. We sample rotations by sampling three Euler angle rotations in the range $\left[0^{\circ},45^{\circ}\right]$ and translations in the range $\left[-0.5,0.5\right]$ on each axis. The rigid transformation is applied to $\mathbf{X}$, leading to $\mathbf{Y}$, which is opposite to PRNet. The right column in Fig.~\ref{fig:partial-to-partial data} shows some examples. The points in $\mathbf{X}$ and $\mathbf{Y}$ are more decentralized than those generated in the manner of PRNet, which means that the overlapping ratio is small in some cases. As a result, it is more difficult to register with this data.


\begin{table}[t]
    \small
    \centering
    \resizebox{\linewidth}{!}{%
        \begin{tabular}{lcccccc}
            \toprule
            \multicolumn{1}{c}{} & \multicolumn{2}{c}{RMSE(\textbf{R})} & \multicolumn{2}{c}{MAE(\textbf{R})} & \multicolumn{2}{c}{Error(\textbf{R})}
            \\ 
            \cmidrule(lr){2-3} \cmidrule(lr){4-5} \cmidrule(lr){6-7}
            \multicolumn{1}{c}{\multirow{-2}{*}{Method}}&\emph{OS} &\emph{TS} &\emph{OS} &\emph{TS} &\emph{OS} &\emph{TS}
            \\ 
            \midrule
            RPMNet~\cite{yew2020-RPMNet} & 0.312 & 6.531 & 0.200 & 2.972 & 0.432 & 8.454 \\
            PRNet~\cite{wang2019prnet} & 5.979 & 13.773 & 3.779 & 9.670 & 7.714 & 20.692 \\
            IDAM~\cite{idam} & 1.605 & 7.725 & 0.905 & 4.364 & 1.850 & 10.940 \\ 
            Ours & 0.766 & 6.258 & 0.347 & 2.877 & 0.873 & 8.606 \\ 
            \bottomrule
        \end{tabular}%
    }
    \caption{Results on 8 axisymmetrical categories in ModelNet40. \emph{OS} and \emph{TS} denotes the results on the once-sampled and our twice-sampled data separately. The performances of all methods are decreasing when changing the data sampling manner from \emph{OS} to \emph{TS}.}
    \label{tab:over-fitting}
\end{table}

\section{More Experiments on ModelNet40}
\label{sec:3}
In this section, we provide more experimental results on ModelNet40~\cite{wu20153d} dataset to validate the robustness and effectiveness of our method. First, we show the decrement of performances between evaluating on the once-sampled (\emph{OS}) and the twice-sampled (\emph{TS}) data in Sec.~\ref{sec:3.1}, which demonstrates the over-fitting issue. Then, we show the results on the test set that generated in the manner of PRNet in Sec.~\ref{sec:3.2}, and the results on the dataset that generated in the manner of RPMNet are shown in Sec.~\ref{sec:3.3}. Besides, the comparison of speed shows the computational efficiency of our method in Sec.~\ref{sec:3.4}. Finally, Sec.~\ref{sec:3.5} explores that how many iterations are needed.

\subsection{Over-fitting Issue}
\label{sec:3.1}
In this experiment, we demonstrate that deep learning (DL) based methods can easily over-fit the original data distribution even with added noises, as shown in Table~\ref{tab:over-fitting}. Note that we train and evaluate on 8 axisymmetrical categories, where point clouds are only rotated at the z-axis. We can find that all partial-to-partial methods can achieve good performances on the \emph{OS} data, however, performances are obviously decreasing when only changing the data sampling manner from \emph{OS} to \emph{TS}. As a result, all the methods can over-fit the distribution of \emph{OS} data, while failing to register the axisymmetrical \emph{TS} data.

\subsection{Results on PRNet dataset}
\label{sec:3.2}
In our main paper, we only show the results on the validation set with Gaussian noise generated in the partial manner of PRNet~\cite{wang2019prnet}. To further demonstrate the robustness of our method, we show results on the unseen categories with Gaussian noise. We add noises sampled from $\mathcal{N}(0,0.01^{2})$ and clipped to $\left[-0.05,0.05\right]$ on each axis, then repeat the second experiment (unseen categories) in our main paper. Table~\ref{tab:PRNet Unseen Categories with Gaussian Noise} shows the performances of various methods. Our method achieves accurate registration and ranks first.

\subsection{Results on RPMNet dataset}
\label{sec:3.3}
In this subsection, we show 4 experimental results on the ModelNet40 dataset with or without Gaussian noise that generated in the data partial manner of RPMNet~\cite{yew2020-RPMNet}.
\begin{table}[t]
    \setlength\tabcolsep{2.5pt} 
    \resizebox{0.48\textwidth}{!}{%
        \begin{tabular}{lcccccccccc}
            \toprule
             &ICP &FGR &PointNetLK &DCP &PRNet &FMR &RPMNet &IDAM &DeepGMR &
            \\ 
            \multirow{-2}{*}{\# points} &\cite{besl1992method} &\cite{zhou2016fast} &\cite{aoki2019pointnetlk} &\cite{wang2019deep} &\cite{wang2019prnet} &\cite{huang2020feature} &\cite{yew2020-RPMNet} &\cite{idam} &\cite{yuan2020deepgmr} &\multirow{-2}{*}{Ours} 
            
            \\ 
            \midrule
            512 &33 &37 &73 &\textcolor{blue}{15} &79 &138 &58 &27 &\textcolor{red}{9} &24
            \\
            1024 &56 &92 &77 &\textcolor{blue}{17} &84 &158 &115 &28 &\textcolor{red}{9} &25
            \\
            2048 &107 &237 &83 &\textcolor{blue}{26} &114 &295 &271 &33 &\textcolor{red}{9} &27 
            \\
            4096 &271 &673 &89 &88 &- &764 &726 &62 &\textcolor{red}{10} &\textcolor{blue}{32} 
            \\ 
            \bottomrule
        \end{tabular}%
    }
    \caption{Speed comparison for registering a point cloud pair of various sizes (in milliseconds). The missing result in the table is due to the limitation in GPU memory.}
    \vspace{-0.37cm}
    \label{tab:inference time}
\end{table}
\vspace{-0.3cm}
\paragraph{Unseen Shapes.}
In this experiment, we train the models on the training set of the first 14 categories and evaluate the registration performances on the validation set of the same categories without noise. Table~\ref{tab:RPMNet}(a) shows the results. 

We can find that all traditional methods and most DL based methods perform poorly because of the large difference in initial positions and partiality. Note that the normals are calculated after the data pre-processing, so that the normals of points in $\mathbf{X}$ can be different from their correspondences in $\mathbf{Y}$. Although Go-ICP~\cite{yang2013go} attempts to improve the performance of the original ICP~\cite{besl1992method} by adopting a brute-force branch-and-bound strategy, it may not suitable for this scene and brings negative implications. Our method outperforms all the traditional and DL based methods except 3 metrics on the \emph{OS} data compared with RPMNet. 

\vspace{-0.4cm}
\paragraph{Unseen Categories.}
We evaluate the performance on unseen categories without noise in this experiment. The models are trained on the first 14 categories and tested on the other 18 categories. The results are summarized in Table~\ref{tab:RPMNet}(b). We can find that the performances of all DL based methods become worse without training on the same categories. Nevertheless, the traditional methods are not affected so much due to the handcrafted features. Our method outperforms all the traditional and DL based methods.

\vspace{-0.4cm}
\paragraph{Gaussian Noise.}
To evaluate the capability of robustness to noise, we add noises sampled from $\mathcal{N}(0,0.01^{2})$ on each axis and clipped to $\left[-0.05,0.05\right]$, then repeat the first two experiments (unseen shapes and unseen categories). Table~\ref{tab:RPMNet}(c)(d) show the performances of different algorithms. FGR~\cite{zhou2016fast} is sensitive to noise, so that it performs much worse than the noise-free case. Almost all the DL based methods become worse with noise injected. Our method achieves the best performance compared to all competitors.


\subsection{Efficiency}
\label{sec:3.4}
We profile the inference times in Table~\ref{tab:inference time}. We test DL based models on a NVIDIA RTX 2080Ti GPU and two 2.10 GHz Intel Xeon Gold 6130 CPUs for the other methods. For our approach, we provide the time of $N=4$ iterations. The computational time is averaged over the entire test set. The speeds of traditional methods are variant under different settings. Note that ICP is accelerated using the k-D tree. We do not compare with Go-ICP because its obviously slow speed. Our method is faster especially with large inputs but is slower than the non-iterative DCP and DeepGMR.
\begin{figure}[t]
\centering
\includegraphics[width=\linewidth]{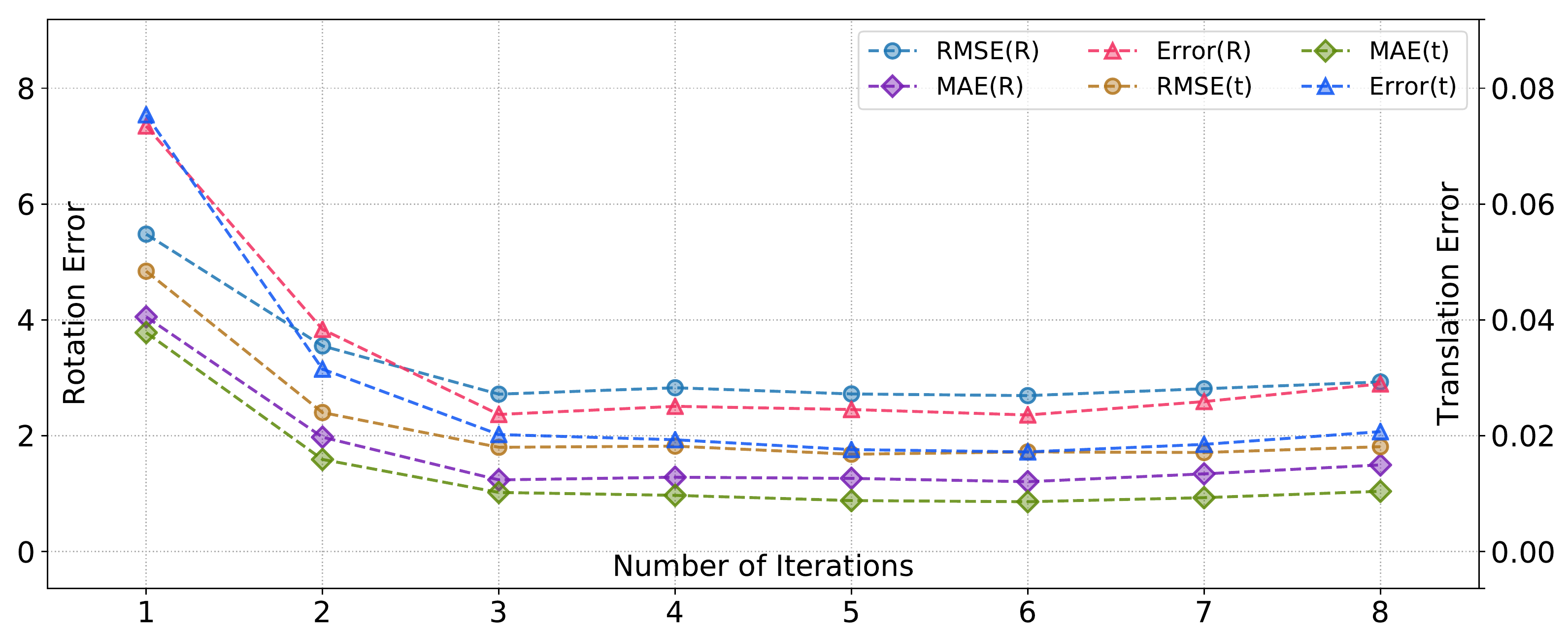}\\
\caption{Anisotropic and isotropic errors of our method over registration iterations.}
\label{fig:number of iteration}
\end{figure}
\subsection{Number of Iterations}
\label{sec:3.5}
The model is trained with different iterations to show that how many iterations are needed. The anisotropic and isotropic errors are calculated after each iteration, as illustrated in Fig.~\ref{fig:number of iteration}. We can find that most performance gains are in the first two iterations, and we choose $N=4$ for the trade-off between the speed and the accuracy in all experiments.
\begin{table*}[t]
    \small
    \resizebox{\linewidth}{!}{%
        \begin{tabular}{lcccccccccccc}
            \hline
            \multicolumn{1}{c}{\multirow{2}{*}{Method}} &\multicolumn{2}{c}{RMSE(\textbf{R})} &\multicolumn{2}{c}{MAE(\textbf{R})} &\multicolumn{2}{c}{RMSE(\textbf{t})} &\multicolumn{2}{c}{MAE(\textbf{t})} &\multicolumn{2}{c}{Error(\textbf{R})} &\multicolumn{2}{c}{Error(\textbf{t})} 
            \\
            \cline{2-13} 
            &\emph{OS} &\emph{TS} &\emph{OS} &\emph{TS} &\emph{OS} &\emph{TS} &\emph{OS} &\emph{TS} &\emph{OS} &\emph{TS} &\emph{OS} &\emph{TS}
            \\ 
            \hline
            ICP~\cite{besl1992method} & 17.439 & 18.588 & 8.954 & 9.628 & 0.0848 & 0.0920 & 0.0460 & 0.0521 & 17.435 & 18.720 & 0.0905 & 0.1026 \\
            Go-ICP~\cite{yang2013go} & 13.081 & 15.214 & 3.617 & 4.650 & 0.0455 & 0.0566 & 0.0169 & 0.0223 & 7.184 & 9.002 & 0.0334 & 0.0445 \\
            Symmetric ICP~\cite{rusinkiewicz2019symmetric} & 6.447 & 7.096 & 5.790 & 6.280 & 0.0615 & 0.0688 & 0.0552 & 0.0617 & 11.340 & 12.531 & 0.1065 & 0.1191 \\
            FGR~\cite{zhou2016fast} & 19.027 & 33.723 & 8.383 & 19.268 & 0.1041 & 0.1593 & 0.0498 & 0.0914 & 15.902 & 35.971 & 0.0981 & 0.1828 \\
            PointNetLK~\cite{aoki2019pointnetlk} & 27.589 & 29.747 & 16.047 & 18.550 & 0.1516 & 0.1841 & 0.0955 & 0.1081 & 30.406 & 32.760 & 0.1907 & 0.1959 \\
            DCP~\cite{wang2019deep} & 7.353 & 7.300 & 4.923 & 4.378 & 0.0657 & 0.0389 & 0.0451 & 0.0272 & 9.624 & 8.853 & 0.0902 & 0.0539 \\
            PRNet~\cite{wang2019prnet} & \textcolor{blue}{3.241} & 5.883 & \textcolor{blue}{1.632} & 3.037 & \textcolor{blue}{0.0181} & 0.0380 & \textcolor{blue}{0.0127} & 0.0237 & \textcolor{blue}{3.095} & 5.974 & \textcolor{blue}{0.0254} & 0.0472 \\
            FMR~\cite{huang2020feature} & 4.819 & \textcolor{blue}{5.304} & 2.488 & \textcolor{blue}{2.779} & 0.0345 & \textcolor{blue}{0.0323} & 0.0158 & \textcolor{blue}{0.0172} & 4.824 & \textcolor{blue}{5.392} & 0.0313 & \textcolor{blue}{0.0342} \\
            IDAM~\cite{idam} & 5.188 & 8.008 & 3.114 & 4.559 & 0.0377 & 0.0484 & 0.0208 & 0.0291 & 5.836 & 8.774 & 0.0418 & 0.0578 \\
            Ours & \textcolor{red}{2.203} & \textcolor{red}{2.563} & \textcolor{red}{0.910} & \textcolor{red}{1.215} & \textcolor{red}{0.0155} & \textcolor{red}{0.0183} & \textcolor{red}{0.0078} & \textcolor{red}{0.0098} & \textcolor{red}{1.809} & \textcolor{red}{2.360} & \textcolor{red}{0.0156} & \textcolor{red}{0.0196} 
            \\ 
            \hline
        \end{tabular}%
    }
    \caption{Results on point clouds of unseen categories with Gaussian noise in ModelNet40, which are generated in the manner of PRNet. Red indicates the best performance and blue indicates the second-best result.}
    \label{tab:PRNet Unseen Categories with Gaussian Noise}
\end{table*}

\begin{table*}[t]
    \small
    \resizebox{\linewidth}{!}{%
        \begin{tabular}{clcccccccccccc}
            \toprule
            & \multicolumn{1}{c}{} & \multicolumn{2}{c}{RMSE(\textbf{R})} & \multicolumn{2}{c}{MAE(\textbf{R})} & \multicolumn{2}{c}{RMSE(\textbf{t})} & \multicolumn{2}{c}{MAE(\textbf{t})} & \multicolumn{2}{c}{Error(\textbf{R})} & \multicolumn{2}{c}{Error(\textbf{t})} 
            \\ 
            \cmidrule(lr){3-4} \cmidrule(lr){5-6} \cmidrule(lr){7-8} \cmidrule(lr){9-10} \cmidrule(lr){11-12} \cmidrule(lr){13-14} 
            &\multicolumn{1}{c}{\multirow{-2}{*}{Method}}&\emph{OS} &\emph{TS} &\emph{OS} &\emph{TS} &\emph{OS} &\emph{TS} &\emph{OS} &\emph{TS} &\emph{OS} &\emph{TS} &\emph{OS} &\emph{TS} \\ 
            \midrule
            & ICP~\cite{besl1992method} & 20.036 & 22.840 & 10.912 & 12.147 & 0.1893 & 0.1931 & 0.1191 & 0.1217 & 22.232 & 24.654 & 0.2597 & 0.2612 \\
            & Go-ICP~\cite{yang2013go} & 70.776 & 71.077 & 39.000 & 38.266 & 0.3111 & 0.3446 & 0.1807 & 0.1936 & 71.597 & 76.492 & 0.3996 & 0.4324 \\
            & Symmetric ICP~\cite{rusinkiewicz2019symmetric} & 10.419 & 11.295 & 8.992 & 9.592 & 0.1367 & 0.1394 & 0.1082 & 0.1124 & 17.954 & 19.571 & 0.2367 & 0.2414 \\
            & FGR~\cite{zhou2016fast} & 48.533 & 46.766 & 29.661 & 29.635 & 0.2920 & 0.3041 & 0.1965 & 0.2078 & 55.855 & 57.685 & 0.4068 & 0.4263 \\
            & PointNetLK~\cite{aoki2019pointnetlk} & 23.866 & 27.482 & 15.070 & 18.627 & 0.2368 & 0.2532 & 0.1623 & 0.1778 & 29.374 & 36.947 & 0.3454 & 0.3691 \\
            & DCP~\cite{wang2019deep} & 12.217 & 11.109 & 9.054 & 8.454 & 0.0695 & 0.0851 & 0.0524 & 0.0599 & 7.835 & 9.216 & 0.1049 & 0.1259 \\
            & RPMNet~\cite{yew2020-RPMNet} & \textcolor{blue}{1.347} & \textcolor{blue}{2.162} & \textcolor{blue}{0.759} & \textcolor{blue}{1.135} & \textcolor{blue}{0.0228} & \textcolor{blue}{0.0267} & \textcolor{blue}{0.0089} & \textcolor{blue}{0.0141} & \textcolor{blue}{1.446} & \textcolor{blue}{2.280} & \textcolor{blue}{0.0193} & \textcolor{blue}{0.0302} \\
            & FMR~\cite{huang2020feature} & 7.642 & 8.033 & 4.823 & 4.999 & 0.1208 & 0.1187 & 0.0723 & 0.0726 & 9.210 & 9.741 & 0.1634 & 0.1617 \\
            & DeepGMR~\cite{yuan2020deepgmr} & 72.310 & 70.886 & 49.769 & 47.853 & 0.3443 & 0.3703 & 0.2462 & 0.2582 & 82.652 & 86.444 & 0.5044 & 0.5354 \\
            \multirow{-10}{*}{\rotatebox{90}{(a) Unseen Shapes}} & Ours & \textcolor{red}{0.771} & \textcolor{red}{1.384} & \textcolor{red}{0.277} & \textcolor{red}{0.542} & \textcolor{red}{0.0154} & \textcolor{red}{0.0226} & \textcolor{red}{0.0056} & \textcolor{red}{0.0093} & \textcolor{red}{0.561} & \textcolor{red}{1.118} & \textcolor{red}{0.0122} & \textcolor{red}{0.0198}  \\ 
            \midrule
            & ICP~\cite{besl1992method} & 20.387 & 22.906 & 12.651 & 13.599 & 0.1887 & 0.1994 & 0.1241 & 0.1286 & 25.085 & 26.819 & 0.2626 & 0.2700 \\
            & Go-ICP~\cite{yang2013go} & 69.747 & 64.455 & 39.646 & 34.017 & 0.3035 & 0.3196 & 0.1788 & 0.1888 & 68.329 & 68.920 & 0.3893 & 0.4091 \\
            & Symmetric ICP~\cite{rusinkiewicz2019symmetric} & 12.291 & 12.333 & 10.841 & 10.746 & 0.1488 & 0.1456 & 0.1212 & 0.1186 & 21.399 & 21.437 & 0.2577 & 0.2521 \\
            & FGR~\cite{zhou2016fast} & 46.161 & 41.644 & 27.475 & 26.193 & 0.2763 & 0.2872 & 0.1818 & 0.1951 & 49.749 & 51.463 & 0.3745 & 0.4003 \\
            & PointNetLK~\cite{aoki2019pointnetlk} & 27.903 & 42.777 & 18.661 & 28.969 & 0.2525 & 0.3210 & 0.1752 & 0.2258 & 36.741 & 53.307 & 0.3671 & 0.4613 \\
            & DCP~\cite{wang2019deep} & 13.387 & 12.507 & 9.971 & 9.414 & 0.0762 & 0.1020 & 0.0570 & 0.0730 & 11.128 & 12.102 & 0.1143 & 0.1493 \\
            & RPMNet~\cite{yew2020-RPMNet} & \textcolor{blue}{3.934} & \textcolor{blue}{7.491} & \textcolor{blue}{1.385} & \textcolor{blue}{2.403} & \textcolor{blue}{0.0441} & \textcolor{blue}{0.0575} & \textcolor{red}{0.0150} & \textcolor{blue}{0.0258} & \textcolor{red}{2.606} & \textcolor{blue}{4.635} & \textcolor{red}{0.0318} & \textcolor{blue}{0.0556} \\
            & FMR~\cite{huang2020feature} & 10.365 & 11.548 & 6.465 & 7.109 & 0.1301 & 0.1330 & 0.0816 & 0.0837 & 12.159 & 13.827 & 0.1773 & 0.1817 \\
            & DeepGMR~\cite{yuan2020deepgmr} & 75.773 & 68.425 & 53.689 & 46.269 & 0.3485 & 0.3667 & 0.2481 & 0.2595 & 85.210 & 87.192 & 0.5074 & 0.5323 \\
            \multirow{-10}{*}{\rotatebox{90}{(b) Unseen Categories}} & Ours & \textcolor{red}{3.719} & \textcolor{red}{4.014} & \textcolor{red}{1.314} & \textcolor{red}{1.619} & \textcolor{red}{0.0392} & \textcolor{red}{0.0406} & \textcolor{blue}{0.0151} & \textcolor{red}{0.0179} & \textcolor{blue}{2.657} & \textcolor{red}{3.206} & \textcolor{blue}{0.0321} & \textcolor{red}{0.0383} \\
            \midrule
            & ICP~\cite{besl1992method} & 20.245 & 23.174 & 11.134 & 12.405 & 0.1902 & 0.1932 & 0.1214 & 0.1231 & 22.580 & 25.147 & 0.2634 & 0.2639 \\
            & Go-ICP~\cite{yang2013go} & 72.221 & 72.030 & 40.516 & 39.308 & 0.3162 & 0.3468 & 0.1860 & 0.1977 & 74.420 & 77.519 & 0.4089 & 0.4405 \\
            & Symmetric ICP~\cite{rusinkiewicz2019symmetric} & 11.087 & 11.731 & 9.671 & 10.042 & 0.1453 & 0.1436 & 0.1157 & 0.1163 & 19.174 & 20.292 & 0.2517 & 0.2486 \\
            & FGR~\cite{zhou2016fast} & 53.186 & 47.816 & 33.189 & 30.572 & 0.3059 & 0.3149 & 0.2117 & 0.2185 & 63.019 & 59.759 & 0.4368 & 0.4459 \\
            & PointNetLK~\cite{aoki2019pointnetlk} & 24.162 & 26.235 & 16.222 & 17.874 & 0.2369 & 0.2582 & 0.1684 & 0.1805 & 32.108 & 36.109 & 0.3555 & 0.3771 \\
            & DCP~\cite{wang2019deep} & 12.387 & 12.393 & 9.147 & 9.534 & 0.0656 & 0.1008 & 0.0495 & 0.0717 & 8.341 & 8.955 & 0.0989 & 0.1516 \\
            & RPMNet~\cite{yew2020-RPMNet} & \textcolor{blue}{1.670} & \textcolor{blue}{2.955} & \textcolor{blue}{0.889} & \textcolor{blue}{1.374} & \textcolor{blue}{0.0310} & \textcolor{blue}{0.0360} & \textcolor{blue}{0.0111} & \textcolor{blue}{0.0163} & \textcolor{blue}{1.692} & \textcolor{blue}{2.746} & \textcolor{blue}{0.0242} & \textcolor{blue}{0.0353} \\
            & FMR~\cite{huang2020feature} & 8.026 & 8.591 & 5.051 & 5.303 & 0.1244 & 0.1249 & 0.0755 & 0.0776 & 9.657 & 10.383 & 0.1702 & 0.1719 \\
            & DeepGMR~\cite{yuan2020deepgmr} & 74.958 & 70.810 & 52.119 & 47.954 & 0.3520 & 0.3689 & 0.2538 & 0.2597 & 86.935 & 87.444 & 0.5189 & 0.5360 \\
            \multirow{-10}{*}{\rotatebox{90}{\makecell[c]{(c) \tabincell{c}{Unseen Shapes\\with Gaussian Noise}}}} & Ours & \textcolor{red}{0.998} & \textcolor{red}{1.522} & \textcolor{red}{0.555} & \textcolor{red}{0.817} & \textcolor{red}{0.0172} & \textcolor{red}{0.0189} & \textcolor{red}{0.0078} & \textcolor{red}{0.0098} & \textcolor{red}{1.078} & \textcolor{red}{1.622} & \textcolor{red}{0.0167} & \textcolor{red}{0.0208} \\
            \midrule
            & ICP~\cite{besl1992method} & 20.566 & 21.893 & 12.786 & 13.402 & 0.1917 & 0.1963 & 0.1265 & 0.1278 & 25.417 & 26.632 & 0.2667 & 0.2679 \\
            & Go-ICP~\cite{yang2013go} & 70.417 & 65.402 & 40.303 & 34.988 & 0.3072 & 0.3233 & 0.1822 & 0.1929 & 69.175 & 71.054 & 0.3962 & 0.4170 \\
            & Symmetric ICP~\cite{rusinkiewicz2019symmetric} & 12.183 & 12.576 & 10.723 & 10.987 & 0.1487 & 0.1478 & 0.1210 & 0.1203 & 21.169 & 21.807 & 0.2576 & 0.2560 \\
            & FGR~\cite{zhou2016fast} & 49.133 & 46.213 & 31.347 & 30.116 & 0.3002 & 0.3034 & 0.2068 & 0.2141 & 56.652 & 58.968 & 0.4230 & 0.4364 \\
            & PointNetLK~\cite{aoki2019pointnetlk} & 26.476 & 29.733 & 19.258 & 21.154 & 0.2542 & 0.2670 & 0.1853 & 0.1937 & 37.688 & 42.027 & 0.3831 & 0.3964 \\
            & DCP~\cite{wang2019deep} & 13.117 & 12.730 & 9.741 & 9.556 & 0.0779 & 0.1072 & 0.0591 & 0.0774 & 11.350 & 12.173 & 0.1187 & 0.1586 \\
            & RPMNet~\cite{yew2020-RPMNet} & \textcolor{blue}{4.118} & \textcolor{blue}{6.160} & \textcolor{blue}{1.589} & \textcolor{blue}{2.467} & \textcolor{blue}{0.0467} & \textcolor{blue}{0.0618} & \textcolor{blue}{0.0175} & \textcolor{blue}{0.0274} & \textcolor{red}{2.983} & \textcolor{blue}{4.913} & \textcolor{blue}{0.0378} & \textcolor{blue}{0.0589} \\
            & FMR~\cite{huang2020feature} & 10.604 & 11.674 & 6.725 & 7.400 & 0.1300 & 0.1364 & 0.0827 & 0.0867 & 12.627 & 14.121 & 0.1788 & 0.1870 \\
            & DeepGMR~\cite{yuan2020deepgmr} & 75.257 & 68.560 & 53.470 & 46.579 & 0.3509 & 0.3735 & 0.2519 & 0.2654 & 84.121 & 87.104 & 0.5180 & 0.5455 \\
            \multirow{-10}{*}{\rotatebox{90}{\makecell[c]{(d) \tabincell{c}{Unseen Categories\\with Gaussian Noise}}}} & Ours & \textcolor{red}{3.572} & \textcolor{red}{4.356} & \textcolor{red}{1.570} & \textcolor{red}{1.924} & \textcolor{red}{0.0391} & \textcolor{red}{0.0486} & \textcolor{red}{0.0172} & \textcolor{red}{0.0223} & \textcolor{blue}{3.073} & \textcolor{red}{3.834} & \textcolor{red}{0.0359} & \textcolor{red}{0.0476} \\ 
            \bottomrule
        \end{tabular}%
    }
    \caption{Results on point clouds of unseen shapes in ModelNet40, which are generated in the manner of RPMNet.}
    \label{tab:RPMNet}
\end{table*}

\begin{figure*}[t]
\centering
\includegraphics[width=\linewidth]{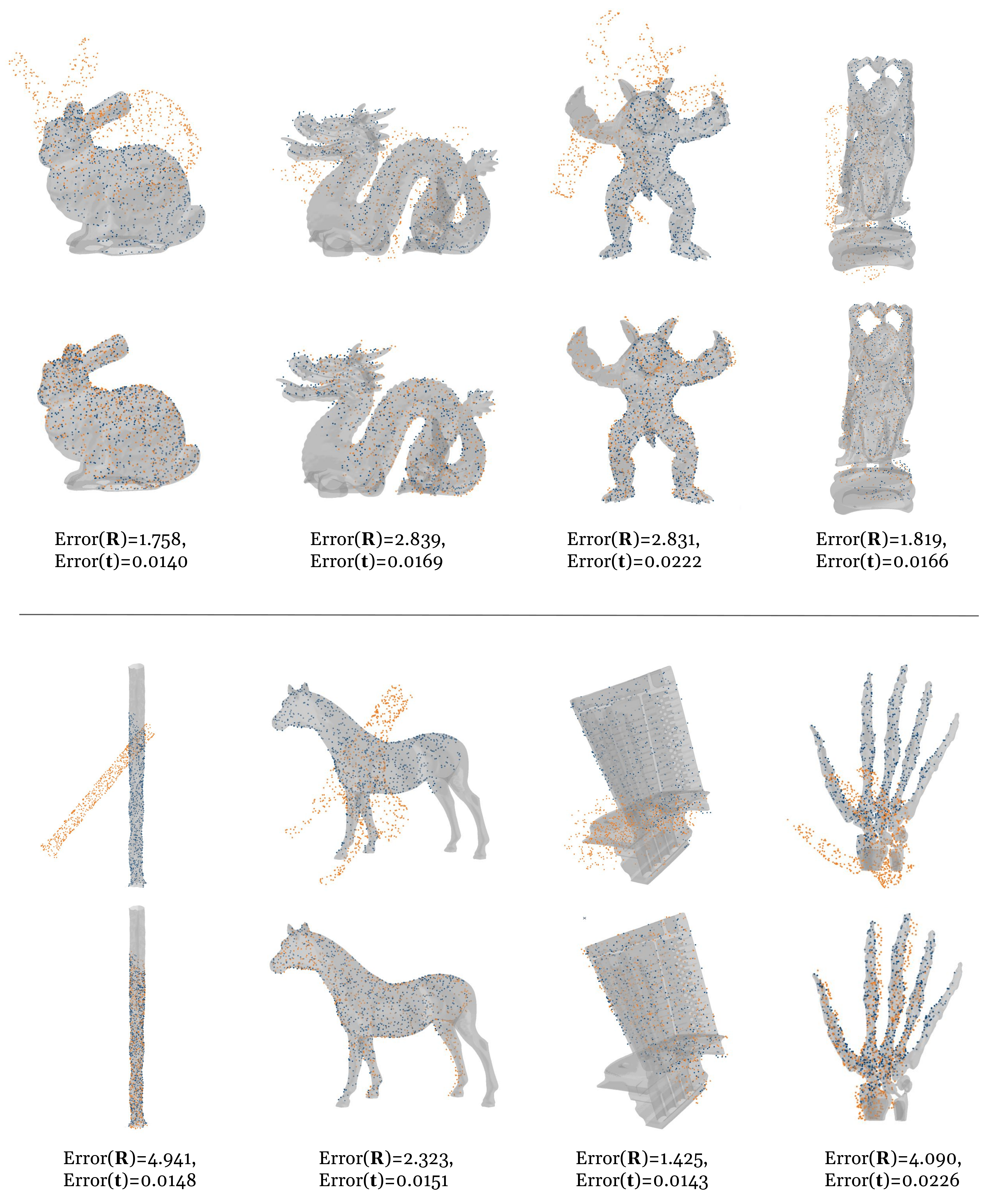}\\
\caption{Results on the Stanford 3D Scan dataset. The model is trained on the training set on ModelNet40, and no fine-tuning is done on Stanford 3D Scan. In each cell separated by the horizontal line, the top row shows the initial positions of the two point clouds, and the bottom row shows the results of registration. Anisotropic and isotropic errors of each result is shown bellow the point clouds.}
\label{fig:results on stanford 3d scan}
\end{figure*}

\end{document}